\documentclass{article}

\PassOptionsToPackage{numbers, compress}{natbib}

    \usepackage[final]{neurips_2024}

\usepackage[utf8]{inputenc} 
\usepackage[T1]{fontenc}    
\usepackage{hyperref}       
\usepackage{url}            
\usepackage{booktabs}       
\usepackage{amsfonts}       
\usepackage{nicefrac}       
\usepackage{microtype}      
\usepackage{xcolor}         
\usepackage{csvsimple}
\usepackage{afterpage}
\usepackage{enumitem}
\usepackage{placeins}
\usepackage{caption}
\usepackage{subcaption}
\usepackage{url}
\usepackage{cleveref}
\usepackage[colorinlistoftodos]{todonotes}  
\usepackage[round-mode=places, round-integer-to-decimal, round-precision=4,
    table-format = 1.4, 
    table-number-alignment=center,
    round-integer-to-decimal]{siunitx}

\title{Using Time-Aware Graph Neural Networks to Predict Temporal Centralities in Dynamic Graphs}

\author{%
  Franziska Heeg \\
  Chair of Machine Learning for Complex Networks\\
  Center for Artificial Intelligence and Data Science (CAIDAS)\\
  Julius-Maximilans-Universität, Würzburg \\
  \texttt{franziska.heeg@uni-wuerzburg.de} \\
   \And
   Ingo Scholtes \\
  Chair of Machine Learning for Complex Networks\\
  Center for Artificial Intelligence and Data Science (CAIDAS)\\
  Julius-Maximilans-Universität, Würzburg \\
  \texttt{ingo.scholtes@uni-wuerzburg.de} \\
}

\begin{document}

\maketitle

\begin{abstract}
  Node centralities play a pivotal role in network science, social network analysis, and recommender systems.
  In temporal data, static path-based centralities like closeness or betweenness can give misleading results about the true importance of nodes in a temporal graph. 
  To address this issue, temporal generalizations of betweenness and closeness have been defined that are based on the shortest time-respecting paths between pairs of nodes.
  However, a major issue of those generalizations is that the calculation of such paths is computationally expensive.
  
  Addressing this issue, we study the application of De Bruijn Graph Neural Networks (DBGNN), a time-aware graph neural network architecture, to predict temporal path-based centralities in time series data.
  We experimentally evaluate our approach in 13 temporal graphs from biological and social systems and show that it considerably improves the prediction of betweenness and closeness centrality compared to (i) a static Graph Convolutional Neural Network, (ii) an efficient sampling-based approximation technique for temporal betweenness, and (iii) two state-of-the-art time-aware graph learning techniques for dynamic graphs.
  \end{abstract}
  
  \maketitle
  
  \section{Motivation}
  \label{sec:motivation}
  
  Node centralities are important in the analysis of complex networks, with applications in network science, social network analysis, and recommender systems.
  An important class of centrality measures are \emph{path-based centralities} like, e.g. betweenness or closeness centrality \cite{bavelas1950communication,freeman1977set}, which are based on the shortest paths between all nodes.
  While centralities in static networks are important, we increasingly have access to time series data on temporal graphs with time-stamped edges.
  Due to the timing and ordering of those edges, the paths in a static time-aggregated representation of such time series data can considerably differ from \emph{time-respecting paths} in the corresponding temporal graph. 
  In a nutshell, two time-stamped edges $(u,v;t)$ and $(v,w;t')$ only form a time-respecting path from node $u$ via $v$ to $w$ iff for the time stamps $t$ and $t'$ we have $t<t'$, i.e. time-respecting paths must minimally respect the arrow of time.
  Moreover, we often consider scenarios where we need to additionally account for a \emph{maximum time difference} $\delta$ between time-stamped edges, i.e. we require $0 < t' - t \leq \delta$ \cite{holme2015modern}.
  Several works have shown that temporal correlations in the sequence of time-stamped edges can significantly change the \emph{causal} topology of a temporal graph, i.e. which nodes can influence each other via time-respecting paths, compared to what is expected based on the static topology \cite{Lentz2013,pan2011path,Pfitzner2013_prl}.
  
  An important consequence of this is that static path-based centralities like closeness or betweenness can give misleading results about the true importance of nodes in temporal graphs. 
  To address this issue, temporal generalizations of betweenness and closeness centrality have been defined that are based on the shortest time-respecting paths between pairs of nodes \cite{tang2010analysing,kim2012temporal,alsayed2015betweenness,tsalouchidou2020temporal}.
  A major issue of those generalizations is that the calculation of time-respecting paths as well as the resulting centralities is computationally expensive \cite{buss2020algorithmic,casteigts2021finding,santoro2022onbra}.
  Addressing this issue, a number of recent works developed methods to approximate temporal betweenness and closeness centralities in temporal graphs \cite{santoro2022onbra}.
  Additionally, few works have used deep (representation) learning techniques to predict computationally expensive path-based centralities in \emph{static} networks \cite{Grando2018,GrandoL15}.
  
\begin{figure*}[]
    \centering
    \includegraphics[width=.65\textwidth]{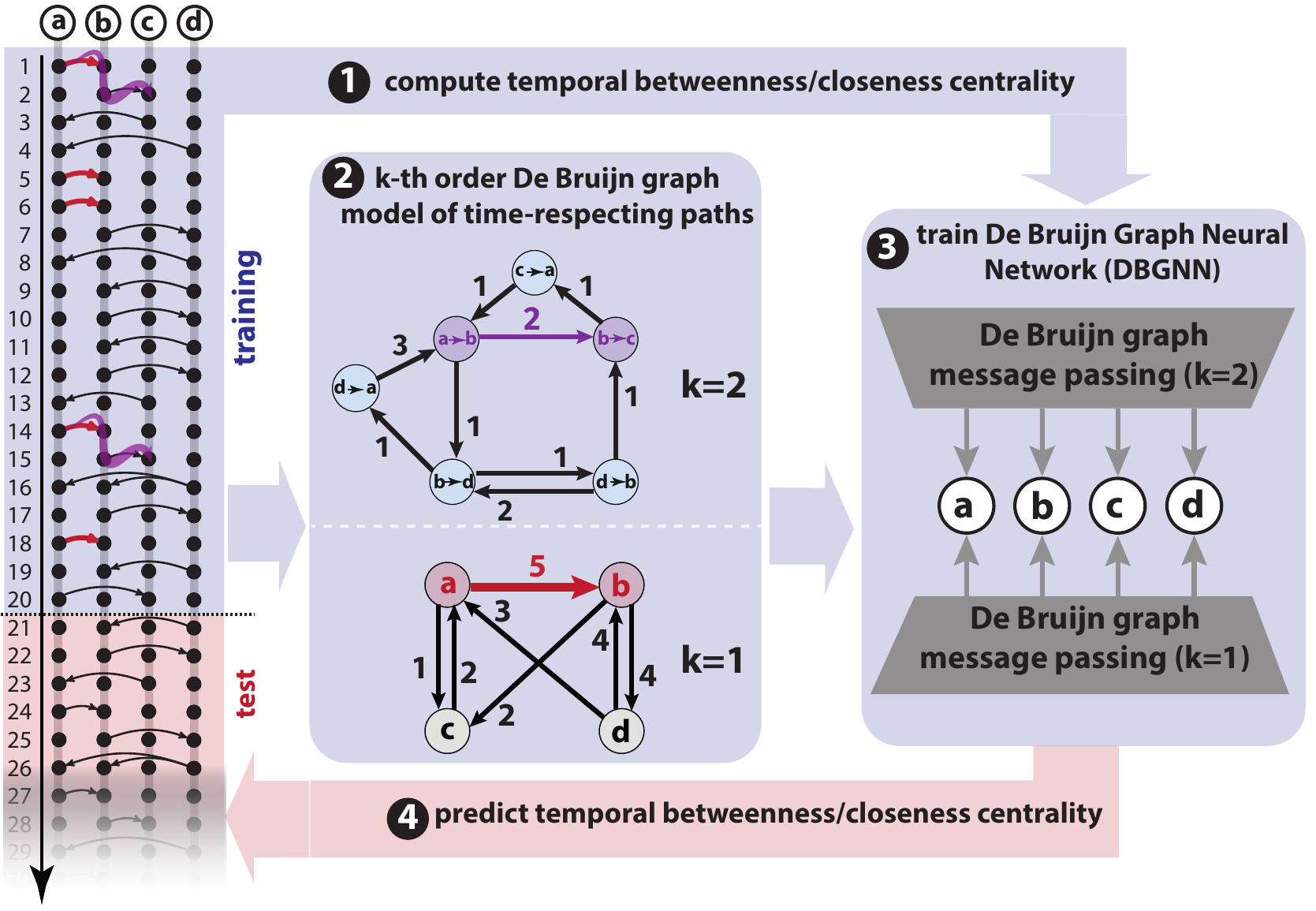}
    \caption{Overview of proposed approach to predict temporal node centralities in a temporal graph: We consider a time-based split in a training and test graph (left). Calculating time-respecting paths in the training split enables us to (1) compute temporal centralities, and (2) fit a $k$-th order De Bruijn graph model for time-respecting paths. The weighted edges in such a $k$-th order De Bruin graph capture frequencies of time-respecting paths of length $k$ (see time-respecting path of length one (red) and two (magenta)). (3) We use these centralities and the k-th order models to train a De Bruijn graph neural network (DBGNN), which allows us to (4) predict temporal centralities in the test graph.}
    \label{fig:intro}
  \end{figure*}

  \paragraph{Research Gap and Contributions}
  To the best of our knowledge, no prior works have considered the application of time-aware graph neural networks to predict path-based centralities in temporal graphs.
  Closing this gap, our work makes the following contributions:
  
  \begin{itemize}[itemsep=0pt,topsep=0pt]
    \item We introduce the problem of predicting temporal betweenness and closeness centralities of nodes in temporal graphs. We consider a situation where we have access to a training graph as well as ground truth temporal centralities and seek to predict the centralities of nodes in a future observation of the same graph, which does not necessarily contain the same node set.
    \item To address this problem, we introduce a deep learning method that utilizes De Bruijn Graph Neural Networks (DBGNN), a recently proposed time-aware graph neural network architecture \cite{qarkaxhija2022bruijn} that is based on higher-order graph models of time-respecting paths, which capture correlations in the sequence of time-stamped edges. An overview of our approach in a toy example of a temporal graph is shown in \Cref{fig:intro}.
    \item We compare our method to a Graph Convolutional Network (GCN), which only considers a static, time-aggregated weighted graph that captures the frequency and topology of edges.
    Evaluating our approach against two deep learning methods for temporal graphs, we consider the time-aware graph embedding method EVO \cite{belth2019remember} as well as the Temporal Graph Network (TGN) framework \cite{rossi2020temporal}.
    We further compare our method to ONBRA\cite{santoro2022onbra}, which efficiently approximates temporal betweenness centralities of nodes to varying degrees of accuracy.
    \item We experimentally evaluate all models in 13 temporal graphs from biological and social systems. Our results show that the application of the time-aware DBGNN architecture considerable improves the prediction of both betweenness and closeness centrality compared to other static and time-aware graph learning techniques. Our method outperforms ONBRA for the prediction of temporal betweenness centralities in large datasets.
  \end{itemize}
  
  In summary, we show that predicting temporal centralities is an interesting temporal graph learning problem, which could be included in community benchmarks \cite{huang2023temporal}.
  Moreover, our study highlights the potential of time-aware deep learning architectures for node-level regression tasks in temporal graphs.
  Finally, our results are a promising step towards an approximation of temporal centralities in large data, with potential applications in social network analysis and recommender systems.

  \section{Background and Related Work}
  \label{sec:relatedwork}
  
  In the following, we provide the background of our work. 
  We first introduce temporal graphs and define time-respecting paths. 
  We then cover generalizations of path-based centralities for nodes in temporal graphs.
  We finally discuss prior works that have studied the prediction, or approximation, of path-based centralities both in static and temporal graphs.
  This will motivate the research gap that is addressed by our work.

  \paragraph{Dynamic Graphs and Time-respecting Paths}
  Apart from static graphs $G=(V,E)$ that capture the topology of edges $E \subseteq V \times V$ between nodes $V$, we increasingly have access to time-stamped interactions that can be modelled as \emph{temporal graphs or networks} \cite{casteigts2012time,Holme2012,holme2015modern}.
  We define a temporal graph as $G^\mathcal{T}=(V,E^\mathcal{T})$ where $V$ is the set of nodes and $E^\mathcal{T}\in V\times V\times \mathbb{R}$ is a set of (possibly directed) time-stamped edges, i.e. an edge $(v,w;t)\in E^\mathcal{T}$ describes an interaction from node $v$ to $w$ occurring at time $t$.
  In our work, we assume that interactions are \emph{instantaneous}, i.e. $(v,w;t) \in E^{\mathcal{T}}$ does not imply that $(v,w;t') \in E^{\mathcal{T}}$ for all $t'>t$.
  Hence, we do not specifically consider \emph{growing networks}, where the time-stamp $t$ is the creation of an edge.
  For a temporal network $G^\mathcal{T}=(V,E^\mathcal{T})$ it is common to consider a static, time-aggregated and weighted graph representation $G=(V,E)$, where $(v,w) \in E$ iff $(v,w;t) \in E^{\mathcal{T}}$ for some time stamp $t$ and for the edge weights we define $w(v,w) = |\{ t \in \mathbb{R} : (v,w;t) \in E^{\mathcal{T}} \} |$, i.e. the number of occurrences of time-stamped edges.

  An important difference to the static case is that, in temporal graphs, the temporal ordering of edges determines \emph{time-respecting paths} \cite{kempe2000connectivity,Holme2012,holme2015modern}.
  For temporal graph $G^\mathcal{T}=(V,E^\mathcal{T})$ we define a \emph{time-respecting path} of length $l$ as node sequence  $v_0,\dots,v_l$ such that the following conditions hold:
  \begin{enumerate}
    \item[$(i)$] $\exists \ t_1,\dots,t_l\ :\ (v_{i-1},v_i;t_i) \in E^\mathcal{T}$ for $i =1,\dots ,l$ ;
    \item[$(ii)$] $0< t_i -t_{i-1}\leq \delta$ for some $\delta\in\mathbb{R}$.
  \end{enumerate}
  In contrast to definitions of time-respecting paths that only require interactions to occur in ascending temporal order, i.e. $0<t_i-t_j$ for $j<i$ \cite{kempe2000connectivity,badie2020efficient}, we also impose a maximum ``waiting time'' $\delta$ \cite{pan2011path,Holme2012}.
  This implies that we only consider time-respecting paths where subsequent interactions occur within a time interval that is often defined by the processes that we study on temporal networks \cite{casteigts2021finding,BadieModiri2020}.
  In line with the definition for static networks, we define a \emph{shortest time-respecting path} between nodes $v$ and $w$ as a (not necessarily unique) time-respecting path of length $l$ such that all other time-respecting paths from $v$ to $w$ have length $l'\geq l$.
  In static graphs a shortest path from $v$ to $w$ is necessarily a \emph{simple} path, i.e. a path where no node occurs more than once in the sequence $v_1, \ldots, v_l$.
  This is not necessarily true for shortest time-respecting path, since --due to the maximum waiting time $\delta$-- we may be forced to move between (possibly the same) nodes to continue a time-respecting path.
  Due to the definition of time-respecting paths with limited waiting time $\delta$, we obtain a \emph{temporal-topological} generalization of shortest paths to temporal graphs that accounts for the temporal ordering and timing of interactions.
  We note that other definitions of \emph{fastest paths} only account for temporal rather than topological distance~\cite{pan2011path}, which we however do not consider in our work.
  
  The definition of time-respecting paths above has the important consequence that the connectivity of nodes via time-respecting paths in a temporal network can be considerably different from paths in the corresponding time-aggregated static network.
  As an example, for a temporal network with two time-stamped edges $(u,v;t)$ and $(v,w;t')$ the time-aggregated network contains a path from $u$ via $v$ to $w$, while a time-respecting path from $u$ via $v$ to $w$ can only exist iff $0 < t'-t \leq \delta$.
  In other words, while connectivity in static graphs is \emph{transitive}, i.e. the existence of edges (or paths) connecting $u$ to $v$ and $v$ to $w$ implies that there exists a path that transitively connects $u$ to $w$, the same does not hold for time-respecting paths.
  A large number of works have shown that this difference between paths in temporal and static graphs influences connectivity and reachability \cite{Lentz2013}, the evolution of dynamical processes like diffusion or epidemic spreading~\cite{rosvall2014memory,Pfitzner2013_prl,vanhems2013estimating,Scholtes2014_natcomm}, cluster patterns \cite{rosvall2014memory,salnikov2016using,lambiotte2019networks}, as well as the controllability of dynamical processes \cite{posfai2014structural}.
  
  \paragraph{Temporal Centralities}
  Another interesting question is how the time dimension of temporal graphs influences the importance or \emph{centrality} of nodes \cite{kim2012temporal}.
  To this end, several works have generalized centrality measures originally defined for static graphs to temporal networks.
  For our purpose we limit ourselves to generalizations of betweenness and closeness centrality, which are defined based on the shortest paths between nodes.
  In a static network, a node $v$ has high \emph{betweenness centrality} if there are many shortest paths that pass through $v$ \cite{freeman1977set} and it has high \emph{closeness centrality} if the overall distance to all other nodes is small \cite{bavelas1950communication}.
  We omit those standard definitions here due to space constraints but include them in \cref{sec:appendix:static}.

  Analogously to betweenness centrality for static graphs, for a temporal graph $G=(V,E^{T})$ we define the \emph{temporal betweenness centrality} of node $v\in V$ as
  \[c_B^{temp} (v)= \sum_{s\neq v\neq t \in V}\frac{\sigma_{s,t}(v)}{\sigma_{s,t}}\]
  where $\sigma_{s,t}$ is the number of shortest \emph{time-respecting} paths from node $s$ to $t$.

  To calculate \emph{temporal closeness centrality} we define the temporal distance $d(u,v)$ between two nodes $u,v\in V$ as the length of a shortest time-respecting paths from $u$ to $v$ and thus obtain 
  \[c_C^{temp}(v) = \frac{1}{\sum_{u\in V}d(u,v)} .\]
  Even though the definitions above closely follow those for static networks, it has been shown that the temporal centralities of nodes can differ considerably from their counterparts in static time-aggregated networks \cite{kim2012temporal,lambiotte2019networks}. 
  These findings highlight the importance of a \emph{time-aware} network analysis, which consider both the timing and temporal ordering of links in temporal graphs.

  \paragraph{Approximating Path-based Centralities}
  While path-based centralities have become an important tool in network analysis, a major issue is the computational complexity of the underlying all-pairs shortest path calculation in large graphs.
  For static networks, this issue can be partially alleviated by smart algorithms that speed up the calculation of betweenness centralities \cite{brandes2001faster}.
  Even with these algorithms, calculating path-based centralities in large graphs is a challenge. 
  Hence, a number of works considered approaches to calculate fast approximations, e.g. based on a random sampling of paths \cite{riondato2014fast,bader2007approximating,haghir2013efficient}.
  Another line of studies either used standard, i.e. not graph-based, machine learning techniques to leverage correlations between different centrality scores \cite{GrandoL15,Grando2018}, or used neural graph embeddings in synthetic scale-free networks to approximate the ranking of nodes \cite{MendoncaBZ21}.
  
  Existing works on the approximation of path-based node centralities in time series data have generally focussed on a fast updating of \emph{static} centralities in \emph{evolving graphs} where edges are added or deleted \cite{Bergamini2015,riondato2018abra}, rather than considering \emph{temporal node centralities}.
  For the calculation of temporal closeness or betweenness centralities, the need to calculate shortest \emph{time-respecting paths} between all pairs of nodes is computationally challenging:
  Temporal closeness centrality minimally requires the traversal of all time-stamped edges for all nodes in the graph, which has a time complexity in $O(n \cdot m)$ where $n$ is the number of nodes and $m$ is the number of time-stamped edges in the temporal graph.
  Building on Brandes' algorithm for static betweenness centrality \cite{brandes2001faster}, a fast algorithm for temporal betweenness centrality with complexity $O(n\cdot m \cdot T)$ (where $T$ is the number of different time stamps in the temporal graph) has recently been proposed in \cite{Buss2024}.
  Considering the approximate estimation of temporal betweenness and closeness centrality in temporal graphs, \cite{Scholtes2016_epjb} generalizes static centralities to higher-order De Bruijn graphs, which capture the time-respecting path structure of a temporal graph.
  \cite{santoro2022onbra} recently proposed a sampling-based estimation of temporal betweenness centralities.
  To the best of our knowledge, no prior works have considered the application of deep graph learning to predict temporal node centralities in temporal graphs, which is the gap addressed by our work.

  \section{A Time-Aware GNN to Predict Temporal Centralities}
  \label{sec:method}
  
  Here, we first present higher-order De Bruijn graph models for time-respecting paths in temporal networks.
  We then describe our deep learning architecture to predict temporal centralities.
  
  \paragraph{Higher-Order De Bruijn Graph Models of Time-respecting paths}
  Each time-respecting path gives rise to an ordered sequence $v_0, v_1, \ldots, v_l$ of traversed nodes.
  Let us consider a $k$-th order Markov chain model, where $P(v_{i}|v_{i-k}, \dots, v_{i-1})$ is the probability that a time-respecting path continues to node $v_{i}$, conditional on the $k$ previously traversed nodes.
  A first-order Markov chain model can be defined based on the frequencies of edges (i.e. paths of length $k=1$) captured in a weighted time-aggregated graph, where  
  \[ 
  P(v_{i}|v_{i-1}) := \frac{w(v_{i-1}, v_i)}{\sum_{j}w(v_{i-1}, v_j)}.
  \]
  While a first-order model is justified if the temporal graph exhibits no patterns in the temporal ordering of time-stamped edges, several works have shown that empirical data exhibit patterns that require higher-order Markov models for time-respecting paths \cite{rosvall2014memory,Scholtes2014_natcomm,salnikov2016using}.
  To address this, for $k>1$ we define a $k$-th order Markov chain model based on frequencies of time-respecting paths of length $k$ as 
  \[ 
  P(v_{i}|v_{i-k}, \dots, v_{i-1}) = \frac{w(v_{i-k}, \dots, v_i)}{\sum_{j}w(v_{i-k}, \ldots, v_{i-1}, v_j)},
  \]
  where $w(v_{0}, \dots, v_k)$ counts the number of time-respecting path $v_{0}, \dots, v_k$ in the underlying temporal graph.
  For a temporal graph $G^\mathcal{T}=(V,E^\mathcal{T})$, this approach defines a static \emph{$k$-th order De Bruijn graph model} $G^{(k)}= (V^{(k)},E^{(k)})$ with 
  \begin{itemize}
    \item $V^{(k)}= \{(v_0,\dots,v_{k-1})\ |\ v_0,\dots,v_{k-1}$ is a time-respecting walk of length $k-1$ in  $G^\mathcal{T}\}$
    \item $(u,v)\in E^{(k)}$ iff
      \subitem $(i)\ \ v = (v_1,\dots, v_k)$ with $v_i=u_i$ for $i=1,\dots,k-1$
      \subitem $(ii)\ u \bigoplus v =  (u_0,\dots,u_{k-1},v_k)$ is a time-respecting path of length $k$ in $G^{\mathcal{T}}$.
  \end{itemize}
  We call this $k$-th order model a \emph{De Bruijn graph model} of time-respecting paths, since it is a generalization of a $k$-dimensional De Bruijn graph \cite{DeBruijn1946}, with the additional constraint that an edge only exists iff the underlying temporal network has a corresponding time-respecting path.
  For $k=1$ the first-order De Bruijn graph corresponds to the commonly used static, time-aggregated graph $G=(V,E)$ of a temporal graph $G^{T}$, where edge can be considered time-respecting paths of length one and which neglects information on time dimension.
  For $k>1$ we obtain \emph{static but time-aware higher-order generalizations of time-aggregated graphs}, which are sensitive to the timing and ordering of time-stamped edges.
  Each node in such a $k$-th order De Bruin graph represents a time-respecting path of length $k-1$, while edges represent time-respecting paths of length $k$.
  Edge weights correspond to the number of observations of time-respecting paths of length $k$ (cf. \cref{fig:intro}).

  \paragraph{De Bruijn Graph Neural Networks for Temporal Centrality Prediction}
  Our approach to predict temporal betweenness and closeness centrality uses the recently proposed De Bruijn Graph Neural Networks (DBGNN), a deep learning architecture that builds on $k$-th order De Bruijn graphs \cite{qarkaxhija2022bruijn}.
  The intuition behind this approach is that, by using message passing in multiple (static) $k$-th order De Bruijn graph models of time-respecting paths, we obtain a \emph{time-aware learning algorithm} that considers both the graph topology as well as the temporal ordering and timing of interactions.
  
  Our proposed method is summarized in \cref{fig:intro}.
  Considering time series data on a temporal graph, we first perform a time-based split of the data into a training and test graph.
  We then calculate temporal closeness and betweenness centralities of nodes in the training graph and consider a supervised node-level regression problem, i.e. we use temporal centralities of nodes in the training graph to train a DBGNN model. 
  To this end, we construct $k$-th order De Bruijn graph models for multiple orders $k$, based on the statistics of time-respecting paths of lengths $k$.
  The maximum order is determined by the temporal correlation length (i.e. the Markov order) present in a temporal graph and can be determined by statistical model selection techniques \cite{Scholtes2017}.
  
  Using the update rule defined in Eq. (1) of \cite{qarkaxhija2022bruijn}, we simultaneously perform message passing in all $k$-th order De Bruijn graphs.
  For each $k$-th order De Bruijn graph this yields a (hidden) representation of $k$-th order nodes.
  To aggregate the resulting representation to actual (first-order) nodes in the temporal graph, we perform message passing in an additional bipartite graph, where each $k$-th order node $(v_0, \ldots, v_{k-1})$ is connected to first-order node $v_{k-1}$ (cf. Eq (2) in \cite{qarkaxhija2022bruijn} and \cref{fig:intro}).
  Taking a node regression perspective, we use a final dense linear layer with a single output.
  We use the trained model to predict the temporal centralities of nodes in the test graph.
  Since the subset of nodes and edges that are active in the training and test graph can differ, our model must be able to generalize to temporal graphs with different nodes as well as different graph topologies.
  To address this, we train our models in an inductive fashion by choosing a suitably large number of dimensions for the one-hot encodings during the training phase.

  Compared to \cite{qarkaxhija2022bruijn}, we introduce two significant technical advances: first, we adapt DBGNN for a node-level regression task, which, to our knowledge, has not been previously explored. 
  Second, we implement a different training procedure designed for forecasting. Unlike the node classification task in \cite{qarkaxhija2022bruijn}, our approach involves training a model on a temporal graph within a training window and subsequently refitting this model to forecast temporal centralities in a future observation, which may include previously unseen nodes and edges. 
  This approach enables our model to generalize to forecasting scenarios involving previously unobserved graph elements, potentially extending its utility to other temporal graph forecasting tasks.

  The implementation of our method is based on the Open Source temporal graph learning library pathpyG\footnote{see \url{https://www.pathpy.net} and \url{https://github.com/pathpy/pathpyG}}. 
  The code of our experiments has been permanently archived at Zenodo\footnote{\url{https://doi.org/10.5281/zenodo.10202791}}.

  \section{Experimental Results}
  \label{sec:results}
  
  With our experimental evaluation we seek to answer the following five research question:
  \begin{itemize}[itemsep=-1.5pt,topsep=0pt]
    \item[\bfseries RQ1] How does the predictive power of a time-aware DBGNN model compare to that of a standard GCN that only uses the static topology and ignores the time dimension of dynamic graphs?
    \item[\bfseries RQ2] How does the performance of the DBGNN model compare to (i) a two step approach that combines the temporal graph embedding EVO \cite{belth2019remember} with a feed-forward neural network, and (ii) TGN \cite{rossi2020temporal}, an end-to-end temporal GNN architecture that does, however, not explicitly consider time-respecting paths.

    \item[\bfseries RQ3]How do the  predictions of the DBGNN architecture compare to the results of ONBRA, a method that aims to approximate temporal betweenness centralities?

    \item[\bfseries RQ4] Which speed-up does our prediction method offer compared to the calculation of temporal node centralities?
    \item[\bfseries RQ5] Does the DBGNN architecture generate node embeddings that facilitate interpretability?
  \end{itemize}
  \paragraph{Experimental setup} 
  We experimentally evaluate the performance of the DBGNN architecture by predicting temporal centralities in 13 empirical temporal graphs.
  We split each temporal graph in training and test graphs, where the training and test graph contain half of the data each.
  Since a maximum order detection in those data sets yields a maximum of two (see \cref{tab:order} in \cref{sec:appendix:additional}), we limit the DBGNN architecture to $k=2$.
  To calculate edge weights of the DBGNN model, we count time-stamped edges as well as time-respecting paths of length two for weights of the first and second-order De Bruijn graph, respectively (cf. \cref{fig:intro}).
  Adopting the approach in \cite{qarkaxhija2022bruijn} we use one message passing layer with 16 hidden dimensions for each order $k$ and one bipartite message passing layer with 8 hidden dimensions.
  We use a sigmoid activation function for higher-order layers and an Exponential Linear Unit (ELU) activation function for the bipartite layer.
  
  As a first time-neglecting baseline model, we use a Graph Convolutional Neural Network (GCN) \cite{KipfW17}, which we apply to the weighted time-aggregated representation of the temporal graphs.
  For the GCN model, we use two message passing layers with 16 and 8 hidden dimensions and a sigmoid activation function, respectively. 
  As input features, we use a one-hot encoding (OHE) of nodes for both architectures. 
  In the case of the DBGNN architecture we apply OHE to nodes in all (higher-order) layers.
  Addressing a node regression task, we use a final dense linear layer with a single output and an ELU activation function, and use mean squared error (MSE) as loss function for both architectures.
  We train both models based on (ground-truth) temporal centralities in the training data, using 1000 epochs with an ADAM optimizer, different learning rates, and weight decay of $5 \cdot 10^{-4}$.
  We tested the use of dropout layers for both architectures, but found the results to be worse.
  In \cref{tab:architecture} and \cref{tab:architecture2} in the appendix we summarize the architecture and report all hyperparameters for both models.
  
  As a second baseline method we use the time-aware graph embedding EVO \cite{belth2019remember}, which models correlations in the sequence of nodes traversed by time-respecting paths.
  Similar to DBGNN, EVO uses these correlations to produce a single static embedding of nodes that captures both the topology and temporal patterns in the dynamic graph.
  Different from DBGNN, EVO does not yield an end-to-end centrality prediction approach, i.e. it only produces node embeddings that can then be used for downstream learning tasks.
  To address centrality prediction, we use 16-dimensional node embeddings produced by EVO for time-respecting paths up to length two.
  We then train a two-layer feed-forward network with a 16-dimensional input, a hidden layer with eight dimensions, and a single output. 
  We use a Rectified Linear Unit (ReLU) activation function for both the hidden layer and the output layer.
  
  As a third baseline, we use the Temporal Graph Networks (TGN) framework, a recently proposed graph neural network architecture for temporal graphs \cite{rossi2020temporal}. Different from DBGNN and EVO the TGN framework does not explicitly consider time-respecting paths but produces a time evolving node embedding that accounts for the sequentiality of interactions and node-wise events.
  To this end, TGN splits a temporal graph into multiple equally-sized batches of consecutive time-stamped interactions. In each of these batches a message passing algorithm is used to update node representations based on time-stamped edges in the current batch as well as a memory of node representations and messages in previous batches. 
  Within each batch the learnable parameters of a TGN model can be trained using a variety of graph learning tasks such as, e.g., link prediction and node classification.

  On the one hand we want batches to be small to obtain a sufficient numbers of batches that we can use to train the model on a given dataset. On the other hand small batch sizes introduce the problem that, due to the small number of time-respecting paths, we cannot calculate meaningful centralities. 
  To address this issue, we calculate the temporal centralities of nodes in a given batch $i$ based on a temporal graph obtained by the batches $i-k+1$ to $i$ for some $k$.
  Adopting this approach we train the TGN model based on the training splits of our data. We then use the trained model to perform a per-batch prediction of temporal centralities for all batches in the test split of our data.
  For TGN, we chose the training and test set such that both contain the same number of batches.
  Finally, we average the prediction scores of the test batches to evaluate the performance of the model.

   In addition to the deep learning methods above, as a final baseline we include ONBRA, a recently proposed sampling-based method, which can estimate temporal betweenness centralities with varying degrees of fidelity by sampling pairs of nodes and calculating shortest temporal paths between them\cite{santoro2022onbra}. By choosing a suitable number of samples, we experimentally adjusted the estimation fidelity of ONBRA, such that the estimation algorithm took approximately the same time as our model. In particular, using the publicly available implementation of the authors\footnote{\url{https://github.com/iliesarpe/ONBRA}}, we estimate temporal betweenness for shortest $\delta$-restless walks for ten iterations and adjust the number of sampled node pairs. In order to get a meaningful comparison, we only run ONBRA on the test window, on which we predicted the temporal betweenness centralities with the other models.

  \paragraph{Data sets} We use 13 data sets on temporal graphs from different contexts, including human contact patterns based on (undirected) proximity or face-to-face relations, time-stamped (directed) E-Mail communication networks, as well as antenna interactions between ants in a colony. 
  An overview of the data sets along with a short description, key characteristics and the source is given in \cref{tab:data} in the appendix.
  All data are publicly available from netzschleuder \cite{netzschleuder} and SNAP \cite{snapnets}.

  \paragraph{Evaluation procedure}
  To evaluate our models, we first fit the pre-trained models to the test graph, i.e. we apply the trained models to the test graph and the trained DBGNN model to the De Bruijn graphs for the test data.
  We then use the trained models to predict temporal closeness and betweenness centralities and compare those predictions to ground truth centralities. 
  For the calculation of temporal closeness centrality, we calculate shortest time-respecting path distances for a given maximum time difference $\delta$ between all pairs of nodes using a variation of Dijkstra's algorithm that traverses all time-stamped edges.
  For temporal betweenness centrality, we adopt a variation of the algorithm proposed in \cite{Buss2024}, which we adjusted to account for shortest time-respecting paths with a maximum time difference $\delta$.
  We will make our code of the temporal centrality calculation available upon acceptance of the manuscript.
  \Cref{fig:intro} provides an illustration of our evaluation approach.
  We use Kendall-Tau and Spearman rank correlation to compare a node ranking based on predicted centralities with a ranking obtained from ground truth centralities.
  Since both rank correlation measures yielded qualitatively similar results, we only report the Spearman correlation.
  Since centrality scores are often used to identify a small set of most central nodes, we further calculate the number of hits in the set of nodes with the top ten predicted centralities.
  Since we repeated each experiment $20$ times, we report the mean and the standard deviation of all scores.
  We repeated all experiments for different learning rates between $0.1$ and $0.0001$ and report the best mean scores.
  The associated learning rates as well as all other hyperparameters of the models are reported in the appendix.
  
  \paragraph{Discussion of results}
  The results of our experiments for temporal betweenness and closeness centralities are shown in \cref{tab:results:betweenness} and \cref{tab:results:closeness}, respectively.
  Considering {\bf RQ1}, we find that our time-aware DBGNN-based architecture significantly outperforms a static GCN model for all 13 data sets and for both evaluation metrics for temporal closeness centrality.
  We further observe a large relative increase of the Spearman rank correlation coefficient ranging between 13 \% for ants-2-1 and 241 \% for eu-email-dept4.
  For temporal betweenness centrality, we find that the proposed DBGNN architecture outperforms a GCN-based prediction in terms of Spearman rank correlation for 12 of the 13 data sets, while we observe better performance of the GCN model for a single data set (haggle).
  For the 12 cases where DBGNN outperforms GCN, we find relative increases in Spearman rank correlation between 3 \% (sp-hospital) and 151 \% (ants-2-2).
  For haggle, where a GCN model outperforms a DBGNN-based prediction, the relative increase is 8 \%.
Additionally,  we observe that all methods generally perform better for temporal closeness centrality compared to temporal betweenness centrality. We attribute this to the specific characteristics of those centralities, which are rooted in their definitions. 
  The temporal closeness centrality of a node only depends on the length of shortest time-respecting paths from that node to all other nodes. Moreover temporal closeness centralities likely exhibit strong correlations between neighboring nodes, which specifically favors a prediction based on neural message passing. 
  In contrast, the temporal betweenness centrality of a node is not only influenced by the length of time-respecting paths but also by the specific sequence of traversed nodes. At the same time, depending on the structure of time-respecting paths, two neighboring nodes can have vastly different temporal betweenness centralities. These factors suggest that the prediction of temporal betweenness centrality is a fundamentally more difficult problem than the prediction of temporal closeness centrality.

  \begin{table*}[ht]
    \caption{Results for prediction of temporal betweenness centrality. Reported values are arithmetic mean across 20 runs and we also report the standard deviation. Bold values represent the best result for a given data set and metric.}
      \label{tab:results:betweenness}
      \centering
      \resizebox{1\textwidth}{!}{%
      \begin{tabular}{l|ll|ll|ll|ll}
\toprule
  & \multicolumn{2}{c|}{DBGNN} & \multicolumn{2}{c|}{GCN} & \multicolumn{2}{c|}{EVO} & \multicolumn{2}{c}{TGN} \\
Experiment & Spearmanr & hitsIn10 & Spearmanr & hitsIn10 & Spearmanr & hitsIn10 & Spearmanr & hitsIn10 \\
\midrule
ants-1-1 & {\bfseries {0.636}} ± 0.063 & {\bfseries {6.050}} ± 1.234 & 0.282 ± 0.147 & 2.350 ± 1.424 & 0.404 ± 0.084 & 4.050 ± 0.51 & 0.263 ± 0.056 & 3.400 ± 0.424 \\
ants-1-2 & {\bfseries {0.655}} ± 0.078 & {\bfseries {4.750}} ± 0.91 & 0.498 ± 0.157 & 4.600 ± 1.392 & 0.339 ± 0.312 & 3.150 ± 1.496 & 0.257 ± 0.084 & 3.725 ± 0.411 \\
ants-2-1 & 0.284 ± 0.073 & {\bfseries {3.550}} ± 0.887 & 0.161 ± 0.126 & 2.200 ± 1.196 & 0.096 ± 0.089 & 2.150 ± 0.366 & {\bfseries {0.309}} ± 0.017 & 2.111 ± 0.192 \\
ants-2-2 & 0.466 ± 0.239 & 4.000 ± 1.451 & 0.185 ± 0.287 & 2.100 ± 1.832 & {\bfseries {0.599}} ± 0.042 & {\bfseries {4.400}} ± 1.957 & 0.357 ± 0.088 & 3.050 ± 0.574 \\
eu-email-dept4 & 0.322 ± 0.062 & 3.900 ± 1.252 & -0.047 ± 0.244 & 1.950 ± 1.432 & {\bfseries {0.486}} ± 0.111 & {\bfseries {4.750}} ± 1.293 & 0.292 ± 0.019 & 4.164 ± 0.378 \\
eu-email-dept2 & 0.383 ± 0.071 & 2.900 ± 1.41 & 0.240 ± 0.089 & {\bfseries {4.000}} ± 1.487 & {\bfseries {0.503}} ± 0.094 & 1.550 ± 0.759 & 0.225 ± 0.023 & 3.274 ± 0.348 \\
eu-email-dept3 & {\bfseries {0.532}} ± 0.068 & 5.700 ± 0.979 & 0.408 ± 0.1 & {\bfseries {6.400}} ± 0.883 & 0.504 ± 0.034 & 3.750 ± 1.743 & 0.236 ± 0.096 & 3.151 ± 0.568 \\
sp-workplace & {\bfseries {0.588}} ± 0.065 & {\bfseries {4.350}} ± 1.04 & 0.441 ± 0.103 & 3.450 ± 0.887 & 0.294 ± 0.072 & 3.400 ± 0.94 & 0.077 ± 0.02 & 1.963 ± 0.064 \\
sp-hypertext & {\bfseries {0.839}} ± 0.017 & 6.300 ± 0.865 & 0.786 ± 0.021 & {\bfseries {6.400}} ± 0.503 & 0.622 ± 0.061 & 3.800 ± 0.696 & 0.260 ± 0.048 & 2.574 ± 0.545 \\
sp-hospital & {\bfseries {0.832}} ± 0.03 & {\bfseries {8.000}} ± 1.257 & 0.804 ± 0.041 & 6.950 ± 0.945 & 0.695 ± 0.067 & 4.600 ± 0.681 & 0.522 ± 0.076 & 6.463 ± 0.402 \\
haggle & 0.626 ± 0.023 & 5.650 ± 1.04 & {\bfseries {0.680}} ± 0.003 & {\bfseries {5.850}} ± 0.366 & 0.630 ± 0.102 & 2.250 ± 0.716 & 0.628 ± 0.013 & 3.302 ± 0.191 \\
manufacturing-email & {\bfseries {0.744}} ± 0.106 & {\bfseries {3.750}} ± 1.251 & 0.404 ± 0.14 & 1.700 ± 0.865 & 0.578 ± 0.111 & 1.850 ± 0.489 & 0.320 ± 0.113 & 1.824 ± 0.265 \\
sp-highschool-2013 & {\bfseries {0.661}} ± 0.03 & {\bfseries {3.500}} ± 1.469 & 0.465 ± 0.055 & 0.850 ± 0.875 & 0.267 ± 0.056 & 1.350 ± 0.587 & 0.114 ± 0.007 & 1.224 ± 0.08 \\
\bottomrule
\end{tabular}

      }
    \end{table*}

    \begin{table*}[htb]
      \caption{Results for prediction of temporal closeness centrality. Reported values are arithmetic mean across 20 runs and we also report the standard deviation. Bold values represent the best result for a given data set and metric.}
        \label{tab:results:closeness}
        \centering
        \resizebox{1\textwidth}{!}{%
        \begin{tabular}{l|ll|ll|ll|ll}
\toprule
  & \multicolumn{2}{c|}{DBGNN} & \multicolumn{2}{c|}{GCN} & \multicolumn{2}{c|}{EVO} & \multicolumn{2}{c}{TGN} \\
Experiment & Spearmanr & hitsIn10 & Spearmanr & hitsIn10 & Spearmanr & hitsIn10 & Spearmanr & hitsIn10 \\
\midrule
ants-1-1 & {\bfseries {0.900}} ± 0.017 & 7.300 ± 0.733 & 0.805 ± 0.011 & {\bfseries {7.850}} ± 0.366 & 0.622 ± 0.045 & 3.800 ± 0.523 & 0.283 ± 0.038 & 2.780 ± 0.54 \\
ants-1-2 & {\bfseries {0.944}} ± 0.006 & {\bfseries {6.650}} ± 0.587 & 0.702 ± 0.003 & 6.000 ± 0.0 & 0.339 ± 0.312 & 3.150 ± 1.496 & 0.305 ± 0.03 & 3.000 ± 0.51 \\
ants-2-1 & {\bfseries {0.974}} ± 0.004 & {\bfseries {8.600}} ± 0.503 & 0.861 ± 0.004 & 7.150 ± 0.366 & 0.117 ± 0.025 & 3.150 ± 0.988 & 0.325 ± 0.032 & 3.080 ± 0.46 \\
ants-2-2 & {\bfseries {0.964}} ± 0.004 & {\bfseries {8.050}} ± 0.394 & 0.662 ± 0.025 & 7.300 ± 0.47 & 0.722 ± 0.057 & 5.550 ± 1.356 & 0.373 ± 0.072 & 3.680 ± 0.642 \\
eu-email-dept4 & {\bfseries {0.972}} ± 0.003 & {\bfseries {8.800}} ± 0.41 & 0.285 ± 0.157 & 2.200 ± 2.042 & 0.486 ± 0.111 & 4.750 ± 1.293 & 0.664 ± 0.026 & 5.084 ± 0.566 \\
eu-email-dept2 & {\bfseries {0.968}} ± 0.006 & {\bfseries {8.550}} ± 0.605 & 0.563 ± 0.007 & 3.000 ± 0.0 & 0.503 ± 0.094 & 1.550 ± 0.759 & 0.574 ± 0.096 & 3.519 ± 0.876 \\
eu-email-dept3 & {\bfseries {0.992}} ± 0.001 & {\bfseries {8.850}} ± 0.366 & 0.653 ± 0.009 & 5.000 ± 0.0 & 0.504 ± 0.034 & 3.750 ± 1.743 & 0.465 ± 0.025 & 3.187 ± 0.277 \\
sp-workplace & {\bfseries {0.893}} ± 0.009 & {\bfseries {7.900}} ± 0.553 & 0.639 ± 0.002 & 6.000 ± 0.0 & 0.388 ± 0.07 & 3.000 ± 0.649 & 0.164 ± 0.061 & 2.692 ± 0.312 \\
sp-hypertext & {\bfseries {0.977}} ± 0.004 & {\bfseries {7.750}} ± 0.55 & 0.809 ± 0.001 & 7.000 ± 0.0 & 0.622 ± 0.061 & 3.800 ± 0.696 & 0.360 ± 0.037 & 3.022 ± 0.607 \\
sp-hospital & {\bfseries {0.918}} ± 0.006 & {\bfseries {7.850}} ± 0.489 & 0.744 ± 0.002 & 5.300 ± 0.47 & 0.695 ± 0.067 & 4.600 ± 0.681 & 0.509 ± 0.058 & 5.649 ± 0.335 \\
haggle & {\bfseries {0.948}} ± 0.005 & {\bfseries {9.300}} ± 0.47 & 0.393 ± 0.001 & 4.950 ± 0.759 & 0.630 ± 0.102 & 2.250 ± 0.716 & 0.559 ± 0.021 & 3.242 ± 0.331 \\
manufacturing-email & {\bfseries {0.971}} ± 0.002 & {\bfseries {7.900}} ± 0.641 & 0.556 ± 0.004 & 3.750 ± 0.444 & 0.716 ± 0.084 & 2.550 ± 1.146 & 0.496 ± 0.028 & 2.258 ± 0.573 \\
sp-highschool-2013 & {\bfseries {0.925}} ± 0.002 & {\bfseries {7.800}} ± 0.41 & 0.540 ± 0.002 & 2.000 ± 0.0 & 0.276 ± 0.026 & 2.900 ± 0.308 & 0.166 ± 0.041 & 1.776 ± 0.165 \\
\bottomrule
\end{tabular}

        }
      \end{table*}
  
  Considering {\bf RQ2}, in \cref{tab:results:betweenness} and \cref{tab:results:closeness} we observe that our proposed DBGNN-based method outperforms both time-aware graph learning techniques TGN and EVO in the majority of data sets, both for temporal closeness and temporal betweenness.
  For temporal betweenness centrality, DBGNN outperforms EVO in all but four data sets (ants-2-2, eu-email-dept4, eu-email-dept2, haggle).
  For the nine data sets where DBGNN outperforms EVO, we find relative performance increases in Spearman rank correlation of up to 139 \% (sp-highschool-2013).
  Results for temporal closeness are even more pronounced, DBGNN outperforming EVO on all data sets, with performance increases ranging from 35 \% (ants-2-2) to 732 \% (ants-2-1).
  This is likely due to DBGNN providing end-to-end learning based on time-respecting paths, as opposed to the two-step approach where we use EVO embeddings as input to a subsequent neural network.
  
  We further find that DBGNN outperforms TGN in all tested cases except for temporal betweenness in the ants-2-1 data set.
  We attribute this to the fact that DBGNN explicitly models patterns in the sequence of nodes traversed by time-respecting paths, which are the basis for the definition of betweenness and closeness centrality.
  The worse performance of TGN can be explained by the fact that the TGN architecture does not use time-respecting paths for the message passing algorithm.
  This makes it -- despite being a time-aware technique that accounts for the temporal evolution of graphs -- a bad choice for temporal graph learning tasks that depend on time-respecting paths.

Considering {\bf RQ3}, for the ONBRA method to approximate temporal betweenness centrality, we find that (i) our method provides a considerably higher performance in terms of Spearman rank correlation for large data sets, and (ii) generally lower mean absolute error across all data sets. Moreover, ONBRA failed to return results for three data sets where our method shows high performance. In \cref{tab:appendix:onbra} in \cref{sec:appendix:onbra} we report the Spearman rank correlation of the results across all data sets, as well as the MAE scores and the time the model took to calculate the estimated centralities. We chose the samples for the ONBRA algorithm such that the time required for the estimation of the centralities approximately matches the inference time for the DBGNN model.
  
  \begin{table*}[ht]
    \caption{Speed-up of the time required for fitting our pretrained model and inference of temporal closeness and betweenness centrality compared to the time required to calculate temporal closeness and betweenness centrality in the validation set.}
      \label{tab:speedup}
      \centering
      \resizebox{.8\textwidth}{!}{%

\begin{tabular}{l|rrr|rrr}
\toprule
 Experiment & \multicolumn{3}{c|}{Closeness} & \multicolumn{3}{c}{Betweenness} \\
 & Fitting+Inference & Centrality & Speed-Up & Fitting+Inference & Centrality & Speed-Up \\
\midrule
ants-1-1 & 0.019 & 0.068 & 3.478 & 0.019 & 0.288 & 14.865 \\
ants-1-2 & 0.012 & 0.056 & 4.510 & 0.012 & 0.107 & 8.721 \\
ants-2-1 & 0.007 & 0.029 & 4.127 & 0.007 & 0.045 & 6.376 \\
ants-2-2 & 0.010 & 0.055 & 5.352 & 0.010 & 0.116 & 11.137 \\
eu-email-dept4 & 0.066 & 1.201 & 18.175 & 0.066 & 1.476 & 22.369 \\
eu-email-dept2 & 0.080 & 1.500 & 18.720 & 0.080 & 1.883 & 23.502 \\
eu-email-dept3 & 0.017 & 0.191 & 11.420 & 0.018 & 0.309 & 17.274 \\
sp-workplace & 0.058 & 1.577 & 27.299 & 0.058 & 4.961 & 86.154 \\
sp-hypertext & 0.207 & 5.908 & 28.570 & 0.207 & 100.517 & 485.728 \\
sp-hospital & 0.465 & 15.289 & 32.908 & 0.464 & 125.977 & 271.461 \\
haggle & 0.069 & 0.516 & 7.431 & 0.069 & 1.032 & 14.910 \\
manufacturing-email & 0.339 & 4.843 & 14.267 & 0.339 & 5.762 & 16.982 \\
sp-highschool-2013 & 2.086 & 91.147 & 43.701 & 2.085 & 2247.218 & 1077.554 \\
\bottomrule
\end{tabular}
      }
    \end{table*}

  Addressing {\bf RQ4}, a potential advantage of our method is that it facilitates \emph{predictions} of temporal centrality node rankings that are much faster than the actual \emph{calculation} of temporal centralities.
  Highlighting this, in \cref{tab:speedup} we report the time needed (i) to fit our pretrained model to the test data, and (ii) to infer the temporal centrality prediction.
  While our approach requires to fit a $k$-th order De Bruijn graph model in the test data, this procedure only requires to calculate time-respecting paths of exactly length $k$, which is a simpler problem than the calculation of all shortest time-respecting paths.
  We compare the combined time of those two steps to the time required to calculate temporal closeness and betweenness centrality in the test graphs, for which we used the fastest known algorithms mentioned in \cref{sec:relatedwork}.
  The results show that our approach provides speed-up factors ranging from approx. 3.5 to 43.7 for temporal closeness and from approx. 6.4 to 1077 for temporal betweenness.

 The corresponding speed-up tables for GCN, TGN and EVO can be found in \cref{sec:appendix:speedup}. Being a much simpler model, the static GCN model provide a higher speed-up (but considerably worse performance). Similarly, EVO provides higher speed-ups but worse predictions. We finally note that the temporal TGN model yields lower speed-ups than our method, despite giving worse predictions.

  A potential criticism of our method could be that the size of a higher-order De Bruijn graph model can be considerably larger than a first-order graph, possibly making training and inference computationally expensive.
  To address this concern, in \cref{sec:appendix:times} we report both the training and inference times of all models across all 13 data sets.
  We find that both the training and inference times of the DBGNN architecture are actually comparable to those of a static GCN.
  We attribute this to the fact that the DBGNN architecture provides a compact, \emph{static but time-aware} De Bruijn graph representation of potentially large time series, rather than requiring a representation of all time-stamped edges.
  We further find that DBGNN has considerably lower training costs than both TGN or EVO.

  Considering {\bf RQ5}, another aspect of our approach to use a time-aware but \emph{static} graph neural network is that the hidden layer activations yield \emph{static} embeddings that are based on the \emph{causal topology} of temporal graphs. 
  This causal topology is influenced by (i) the topology of links, and (ii) their timing and temporal ordering.
  To explain the favorable performance of our model compared to a static GCN, we hypothesize that nodes for which our model learns similar embeddings also have more similar temporal centralities, compared to embeddings generated by a GCN model.
  To test this, we apply a dimensionality reduction to the node activations generated by the last 8-dimensional bipartite layer in the DBGNN architecture, comparing it to the representation obtained from (i) the last message passing layer of a GCN model and (ii) an EVO embedding.
  In \cref{fig:embedding} in \cref{sec:embeddings} we show the resulting embeddings for one representative prediction of temporal closeness and betweeness in the eu-email-dept4 data, where the color gradient highlights ground truth node centralities in the test data.
  The plot shows that the time-aware DBGNN architecture better captures the ranking of nodes compared to a time-neglecting GCN as well as the EVO embeddings.
  We again attribute this to the end-to-end learning approach provided by DBGNN compared to the two-step approach of EVO.

  \section{Conclusion} 
  \label{sec:conclusion}
  
  In summary, we investigate the problem of predicting temporal betweenness and closeness centralities in temporal graphs. 
  We use a recently proposed time-aware graph neural network architecture, which relies on higher-order De Bruijn graph models of time-respecting paths.
  An empirical study in which we compare our approach with a time-neglecting static graph neural network demonstrates the potential of our method.
  We find that our approach considerably outperforms other time-aware graph learning techniques that (i) either do not consider time-respecting paths, or (ii) do not provide an end-to-end approach where the learning of node representations is integrated with the prediction task.
  A comparative analysis in 13 empirical temporal graphs highlights differences between static and temporal centralities that are likely due to the underlying temporal patterns, and shows that our model is generally better at predicting temporal closeness compared to betweenness.
  A scalability analysis reveals that our prediction approach provides a considerable speed-up compared to the exact calculation of temporal node centralities, yielding speed-up factors between 3.5 and 1077.
  We finally investigate (static) embeddings produced by the last message passing layer of our architecture and show that they better capture temporal centralities compared to GCN.
  \paragraph{Open questions and future work} Our work necessarily leaves open questions that should be addressed in future work.
  Rather than optimizing the predictive performance of our model, the focus of the present work was to highlight the potential of time-aware graph neural networks for temporal centrality prediction.
  We thus have not performed an exhaustive optimization of hyperparameters such as, e.g., the maximum time difference $\delta$, the maximum order $k$ of the De Bruijn graphs used in the DBGNN architecture, or the number and width of graph convolutional layers.
  While we do report optimal values across three learning rates for all models, a more thorough investigation of the influence of those hyperparameters is future work.
  Moreover, we did not utilize additional node features like, e.g., node degrees, static centralities, or node embeddings that could further improve our results.
  Another aspect that we have not studied in our work is the impact of the size of the training data, i.e. how little training data is sufficient to predict temporal centralities with reasonable accuracy, and where the trade-offs in the choice of the training size are.
  An interesting further question is whether our approach could be adapted to support a fully \emph{inductive} setting, i.e. to train our model on a set of dynamic graphs and then use the trained model to predict temporal centralities in other, previously unseen networks.
  Similarly, for some data sets with non-stationary temporal patterns it could be beneficial to train a De Bruijn Graph Neural Network based on a sliding window approach, hence adjusting the model (and predictions) as time progresses.
  Such a combination of the concepts behind TGN and DBGNN could yield better results in a number of practical settings. 
  
  We believe that our work is of high practical relevance for applications of knowledge discovery and machine learning in time-stamped relational data.
  For dynamic social network analysis, our approach allows to quickly estimate temporal centralities whose calculation is computationally expensive.
  More generally, our study highlights the potential of \emph{compact, static but time-aware graph neural network architectures} for node-level regression in temporal graphs.
  
  \section*{Acknowledgments}
  IS acknowledges support by the Swiss National Science Foundation (SNF), grant no. 176938 and the German Federal Ministry of Education and Research, grant no. 031L0311A (TissueNet).

  \bibliographystyle{abbrvnat}
  \bibliography{library}
  
  
  \clearpage
  
  
  \appendix  
  
  \section{Details on empirical datasets}
  \begin{table*}[ht]
    \caption{Overview of time series data sets used in the experiment evaluation}
      \label{tab:data}
      \centering
      \resizebox{\textwidth}{!}{%
      \begin{tabular}{l l c|r r r c c }%
      \toprule
      data set & Description & Ref & Nodes & Edges & Temporal Edges & Directed & $\delta$ \\
      \midrule
      ants-1-1 & Ant Antenna interactions, colony 1 - filming 1&\cite{blonder2011time}&89&947&1,911&True & 30 sec\\ 
      ants-1-2 &Ant Antenna interactions, colony 1 - filming 2&\cite{blonder2011time}&72&862&1,820&True & 30 sec\\ 
      ants-2-1 &Ant Antenna interactions, colony 2 - filming 1&\cite{blonder2011time}&71&636&975&True & 30 sec\\ 
      ants-2-2 &Ant Antenna interactions, colony 2 - filming 2&\cite{blonder2011time}&69&769&1,917&True & 30 sec\\ 
      company-emails & E-Mail exchanges in manufacturing company&\cite{nurek2020combining}&167&5,784&82,927&True & 60 mins\\ 
      eu-email-dept2 & E-Mail exchanges in EU institution (dept 2)& \cite{paranjape2017motifs} &162&1,772&46,772&True & 60 mins \\  
      eu-email-dept3 & E-Mail exchanges in EU institution (dept 3)& \cite{paranjape2017motifs} &89&1,506&12,216&True & 60 mins\\ 
      eu-email-dept4 & E-Mail exchanges in EU institution (dept 4)& \cite{paranjape2017motifs} &142&1,375&48,141&True & 60 mins\\ 
    sp-hospital & Face-to-face interactions in a hospital&\cite{vanhems2013estimating}&75&1,139&32,424&False & 60 mins\\
    sp-hypertext & Face-to-face interactions at conference&\cite{isella2011s}&113&2,498&20,818&False& 60 mins\\ 
    sp-workplace & Face-to-face interactions in a workspace& \cite{genois2018can}&92&755&9,827&False & 60 mins\\ 
    sp-highschool & Face-to-face interactions in a highschool &\cite{mastrandrea2015contact}&327&5,818&188,508&False & 60 mins\\ 
    haggle & Human proximity recorded by smart devices&\cite{chaintreau2007impact}&274&2,899&28,244&False & 1 min\\ 
  
  \bottomrule
      \end{tabular}
      }
    \end{table*}

  \section{Comparison of training and inference times across models}
  \label{sec:appendix:times}

  In the following, we investigate the training and inference times of all models for all of the 13 data sets.
  We specifically compare both the training and the inference times of our DBGNN-based architecture with those of the baseline methods GCN, EVO, and TGN. 
  For EVO, the training time is dominated by the time required to compute embeddings, while the time required to train the subsequent feed-forward neural network is negligible.
  The inference time for EVO is exclusively based on the inference time of the feed-forward neural network.
  The results for betweenness and closeness centrality are shown in \cref{tab:appendix:scalability:betweenness} and \cref{tab:appendix:scalability:closeness}, respectively.
  Both tables show that the computational requirements of the DBGNN and GCN model are comparable, both during training and inference.
  The training of the EVO and TGN-based models requires substantially more time. 
  For EVO, this is due to the computational complexity of the embedding calculation, which requires to simulate random walks for the underlying node2vec embedding \cite{Grover2016_node2vec}.
  For TGN, this is due to the necessity to calculate ground truth centralities separately for each batch used in the per-batch training procedure.

  \begin{table*}[!htbp]
    \caption{Training and inference time for betweenness centrality in seconds}
      \label{tab:appendix:scalability:betweenness}
      \centering
      \resizebox{\textwidth}{!}{%
      \begin{tabular}{l|rrrr|rrrr}
\toprule
 & \multicolumn{4}{c|}{Train} & \multicolumn{4}{c}{Test} \\
 & DBGNN & GCN & EVO & TGN & DBGNN & GCN & EVO & TGN \\
\midrule
ants-1-1 & 6.3023 & 5.1324 & 1.3095 & 35.7600 & 0.0026 & 0.0021 & 0.0037 & 0.1140 \\
ants-1-2 & 6.0507 & 5.1753 & 1.4937 & 31.9830 & 0.0022 & 0.0019 & 0.0038 & 0.1010 \\
ants-2-1 & 6.0295 & 5.1794 & 1.4455 & 4.9620 & 0.0023 & 0.0019 & 0.0040 & 0.0340 \\
ants-2-2 & 6.3356 & 5.2399 & 1.3833 & 17.2250 & 0.0023 & 0.0018 & 0.0037 & 0.0590 \\
eu-email-dept4 & 6.0213 & 5.0822 & 1.2874 & 386.1770 & 0.0024 & 0.0019 & 0.0035 & 0.7480 \\
eu-email-dept2 & 6.0819 & 5.1082 & 1.2877 & 730.1510 & 0.0025 & 0.0019 & 0.0036 & 1.3380 \\
eu-email-dept3 & 6.0829 & 5.2514 & 1.2815 & 129.3010 & 0.0024 & 0.0020 & 0.0033 & 0.3520 \\
sp-workplace & 6.4887 & 5.2195 & 1.4293 & 27.6400 & 0.0030 & 0.0019 & 0.0038 & 0.0960 \\
sp-hypertext & 6.0371 & 5.2276 & 1.2776 & 86.9330 & 0.0027 & 0.0020 & 0.0037 & 0.1970 \\
sp-hospital & 5.9835 & 5.0898 & 1.2724 & 166.2600 & 0.0025 & 0.0020 & 0.0036 & 0.3190 \\
haggle & 6.0887 & 5.1281 & 1.2939 & 110.9990 & 0.0028 & 0.0020 & 0.0036 & 0.2520 \\
manufacturing-email & 6.3259 & 5.2400 & 1.5247 & 567.5910 & 0.0030 & 0.0021 & 0.0036 & 1.0010 \\
sp-highschool-2013 & 6.5781 & 4.7907 & 1.5916 & 1360.5520 & 0.0036 & 0.0020 & 0.0044 & 2.2450 \\
\bottomrule
\end{tabular}

      }
    \end{table*}

    \begin{table*}[ht]
      \caption{Training and inference time for closeness centrality in seconds}
        \label{tab:appendix:scalability:closeness}
        \centering
        \resizebox{\textwidth}{!}{%
        \begin{tabular}{l|rrrr|rrrr}
\toprule
 & \multicolumn{4}{c|}{Train} & \multicolumn{4}{c}{Test} \\
 & DBGNN & GCN & EVO & TGN & DBGNN & GCN & EVO & TGN \\
\midrule
ants-1-1 & 6.3638 & 5.4887 & 1.3095 & 33.7800 & 0.0026 & 0.0021 & 0.0037 & 0.1100 \\
ants-1-2 & 6.2833 & 5.4537 & 1.4937 & 18.9370 & 0.0025 & 0.0021 & 0.0038 & 0.0620 \\
ants-2-1 & 6.3401 & 5.5003 & 1.4455 & 16.7240 & 0.0024 & 0.0020 & 0.0040 & 0.0550 \\
ants-2-2 & 6.3093 & 5.4556 & 1.3833 & 33.8390 & 0.0026 & 0.0020 & 0.0037 & 0.1060 \\
eu-email-dept4 & 6.3398 & 5.4577 & 1.2874 & 694.4370 & 0.0025 & 0.0021 & 0.0035 & 1.3430 \\
eu-email-dept2 & 6.3222 & 5.4565 & 1.2877 & 371.8710 & 0.0027 & 0.0022 & 0.0036 & 0.8190 \\
eu-email-dept3 & 6.3126 & 5.4473 & 1.2815 & 129.2170 & 0.0028 & 0.0022 & 0.0033 & 0.3340 \\
sp-workplace & 6.5001 & 5.5629 & 1.4293 & 52.0420 & 0.0029 & 0.0020 & 0.0038 & 0.1300 \\
sp-hypertext & 6.5624 & 5.5350 & 1.2776 & 94.0060 & 0.0030 & 0.0021 & 0.0037 & 0.2140 \\
sp-hospital & 6.5881 & 5.5314 & 1.2724 & 171.0810 & 0.0032 & 0.0021 & 0.0036 & 0.3350 \\
haggle & 6.6309 & 5.5206 & 1.2939 & 116.0810 & 0.0030 & 0.0021 & 0.0036 & 0.2640 \\
manufacturing-email & 6.5784 & 5.6821 & 1.5247 & 801.6310 & 0.0032 & 0.0022 & 0.0036 & 1.4180 \\
sp-highschool-2013 & 6.5746 & 5.6309 & 1.5916 & 1879.3700 & 0.0030 & 0.0022 & 0.0044 & 3.2890 \\
\bottomrule
\end{tabular}

        }
    \end{table*}

    \section{ONBRA results}
    \label{sec:appendix:onbra}
    The results of the ONBRA model for the betweenness centrality are shown in table \ref{tab:appendix:onbra}

    \begin{table*}[ht]
      \caption{MAE and Spearman rank correlation for ONBRA estimation of betweenness centrality}
        \label{tab:appendix:onbra}
        \centering
        \resizebox{\textwidth}{!}{%
        \begin{tabular}{l|lll}
\toprule
 & MAE & Spearmanr & Time\\
\midrule
ants-1-1 & 261.31289 $\pm$ 0.00151 & 0.95429 $\pm$ 0.0062 & 0.0181 $\pm$ 0.00031 \\
ants-1-2 & 34.95047 $\pm$ 0.00076 & 0.8344 $\pm$ 0.03744 & 0.0111 $\pm$ 0.00011 \\
ants-2-1 & 4.08689 $\pm$ 0.00019 & 0.56815 $\pm$ 0.10243 & 0.0068 $\pm$ 7e-05 \\
ants-2-2 & 45.82778 $\pm$ 0.00011 & 0.99184 $\pm$ 0.00178 & 0.42462 $\pm$ 0.01373 \\
eu-email-dept4 & 4.64783 $\pm$ 0.00011 & 0.39658 $\pm$ 0.05351 & 0.06626 $\pm$ 0.00334 \\
eu-email-dept2 & 3.23876 $\pm$ 0.00012 & NaN & 0.07984 $\pm$ 0.00226 \\
eu-email-dept3 & 2.80843 $\pm$ 0.00013 & 0.39598 $\pm$ 0.06837 & 0.01875 $\pm$ 0.0023 \\
sp-workplace & 94.92572 $\pm$ 0.00498 & 0.48874 $\pm$ 0.13345 & 0.15363 $\pm$ 0.38683 \\
sp-hypertext & NaN & NaN & NaN \\
sp-hospital & NaN & NaN & NaN \\
haggle & 13.73402 $\pm$ 0.00026 & 0.40592 $\pm$ 0.10224 & 0.06889 $\pm$ 0.00925 \\
manufacturing-email & 84.03212 $\pm$ 0.00071 & 0.65181 $\pm$ 0.0414 & 0.24526 $\pm$ 0.02451 \\
sp-highschool-2013 & 831.53583 $\pm$ 0.0007 & 0.48636 $\pm$ 0.0479 & 2.93658 $\pm$ 0.56495 \\
\bottomrule
\end{tabular}

        }
    \end{table*}

  \section{Speed-up}
  \label{sec:appendix:speedup}
Tables \ref{tab:appendix:speedup:gcn}, \ref{tab:appendix:speedup:tgn} and \ref{tab:appendix:speedup:evo} show the speed-ups of the prediction model GCN, TGN and EVO compared to the exact calculation of the node centralities

  \begin{table*}[ht]
    \caption{Speed-up of GCN }
      \label{tab:appendix:speedup:gcn}
      \centering
      \resizebox{\textwidth}{!}{%
      \begin{tabular}{l|rrr|rrr}
\toprule
 & \multicolumn{3}{c|}{Betweenness} & \multicolumn{3}{c}{Closeness} \\
 & Inference time & Centrality & Speed-up & Inference time & Centrality & Speed-up \\
\midrule
ants-1-1 & 0.0021 & 0.2876 & 134.7530 & 0.0021 & 0.0678 & 32.2630 \\
ants-1-2 & 0.0019 & 0.1070 & 55.5940 & 0.0021 & 0.0558 & 26.2340 \\
ants-2-1 & 0.0019 & 0.0450 & 24.3330 & 0.0020 & 0.0293 & 14.6190 \\
ants-2-2 & 0.0018 & 0.1156 & 65.1200 & 0.0020 & 0.0554 & 27.0050 \\
eu-email-dept4 & 0.0019 & 1.4759 & 766.7580 & 0.0021 & 1.2011 & 577.7060 \\
eu-email-dept2 & 0.0019 & 1.8833 & 990.9080 & 0.0022 & 1.5004 & 674.2110 \\
eu-email-dept3 & 0.0020 & 0.3086 & 156.1860 & 0.0022 & 0.1911 & 85.8470 \\
sp-workplace & 0.0019 & 4.9610 & 2576.6020 & 0.0020 & 1.5769 & 807.7110 \\
sp-hypertext & 0.0020 & 100.5170 & 50256.4670 & 0.0021 & 5.9080 & 2778.1840 \\
sp-hospital & 0.0020 & 125.9772 & 62201.1900 & 0.0021 & 15.2888 & 7367.6280 \\
haggle & 0.0020 & 1.0321 & 515.9920 & 0.0021 & 0.5155 & 248.3130 \\
manufacturing-email & 0.0021 & 5.7621 & 2710.2200 & 0.0022 & 4.8435 & 2196.1540 \\
sp-highschool-2013 & 0.0020 & 2247.2180 & 1136946.6840 & 0.0022 & 91.1466 & 41897.8210 \\
\bottomrule
\end{tabular}

      }
  \end{table*}

  \begin{table*}[ht]
    \caption{Speed-up of TGN }
      \label{tab:appendix:speedup:tgn}
      \centering
      \resizebox{\textwidth}{!}{%
      \begin{tabular}{l|rrr|rrr}
\toprule
 & \multicolumn{3}{c|}{Betweenness} & \multicolumn{3}{c}{Closeness} \\
 & Inference time & Centrality & Speed-up & Inference time & Centrality & Speed-up \\
\midrule
ants-1-1 & 0.1140 & 0.2876 & 2.5230 & 0.1100 & 0.0678 & 0.6160 \\
ants-1-2 & 0.1010 & 0.1070 & 1.0600 & 0.0620 & 0.0558 & 0.8990 \\
ants-2-1 & 0.0340 & 0.0450 & 1.3240 & 0.0550 & 0.0293 & 0.5330 \\
ants-2-2 & 0.0590 & 0.1156 & 1.9590 & 0.1060 & 0.0554 & 0.5220 \\
eu-email-dept4 & 0.7480 & 1.4759 & 1.9730 & 1.3430 & 1.2011 & 0.8940 \\
eu-email-dept2 & 1.3380 & 1.8833 & 1.4080 & 0.8190 & 1.5004 & 1.8320 \\
eu-email-dept3 & 0.3520 & 0.3086 & 0.8770 & 0.3340 & 0.1911 & 0.5720 \\
sp-workplace & 0.0960 & 4.9610 & 51.6780 & 0.1300 & 1.5769 & 12.1300 \\
sp-hypertext & 0.1970 & 100.5170 & 510.2390 & 0.2140 & 5.9080 & 27.6080 \\
sp-hospital & 0.3190 & 125.9772 & 394.9130 & 0.3350 & 15.2888 & 45.6380 \\
haggle & 0.2520 & 1.0321 & 4.0960 & 0.2640 & 0.5155 & 1.9530 \\
manufacturing-email & 1.0010 & 5.7621 & 5.7560 & 1.4180 & 4.8435 & 3.4160 \\
sp-highschool-2013 & 2.2450 & 2247.2180 & 1000.9880 & 3.2890 & 91.1466 & 27.7130 \\
\bottomrule
\end{tabular}

      }
  \end{table*}

  \begin{table*}[ht]
    \caption{Speed-up of EVO }
      \label{tab:appendix:speedup:evo}
      \centering
      \resizebox{\textwidth}{!}{%
      \begin{tabular}{l|rrr|rrr}
\toprule
 & \multicolumn{3}{c|}{Betweenness} & \multicolumn{3}{c}{Closeness} \\
 & Inference time & Centrality & Speed-up & Inference time & Centrality & Speed-up \\
\midrule
ants-1-1 & 0.0037 & 0.2876 & 76.8720 & 0.0037 & 0.0678 & 18.1150 \\
ants-1-2 & 0.0038 & 0.1070 & 28.4500 & 0.0038 & 0.0558 & 14.8230 \\
ants-2-1 & 0.0040 & 0.0450 & 11.2570 & 0.0040 & 0.0293 & 7.3250 \\
ants-2-2 & 0.0037 & 0.1156 & 31.1560 & 0.0037 & 0.0554 & 14.9260 \\
eu-email-dept4 & 0.0035 & 1.4759 & 424.8480 & 0.0035 & 1.2011 & 345.7490 \\
eu-email-dept2 & 0.0036 & 1.8833 & 517.6840 & 0.0036 & 1.5004 & 412.4260 \\
eu-email-dept3 & 0.0033 & 0.3086 & 93.6080 & 0.0033 & 0.1911 & 57.9480 \\
sp-workplace & 0.0038 & 4.9610 & 1302.1110 & 0.0038 & 1.5769 & 413.8860 \\
sp-hypertext & 0.0037 & 100.5170 & 26984.4380 & 0.0037 & 5.9080 & 1586.0440 \\
sp-hospital & 0.0036 & 125.9772 & 35100.9190 & 0.0036 & 15.2888 & 4259.9130 \\
haggle & 0.0036 & 1.0321 & 286.0590 & 0.0036 & 0.5155 & 142.8800 \\
manufacturing-email & 0.0036 & 5.7621 & 1594.8320 & 0.0036 & 4.8435 & 1340.5670 \\
sp-highschool-2013 & 0.0044 & 2247.2180 & 514590.7920 & 0.0044 & 91.1466 & 20871.6800 \\
\bottomrule
\end{tabular}

      }
  \end{table*}

  \section{Scalability}
  \label{sec:appendix:scalability}

  The computational complexity of our model is linear in the number of time-respecting paths of length two in the temporal graph. This number can be bounded above by the number of paths of length two in the (static) graph, which can be theoretically bounded by $n \cdot \lambda_{1}^2$, where $n$ is the number of nodes and $\lambda_1$ is the largest eigenvalue of the adjacency matrix of the (undirected) static graph. We note that for the fully connected graph with the special case of $\lambda_1 = n$ we obtain an upper bound $n^3$ for the number of time-respecting paths of length two. This corresponds to all length three sequences of $n$ nodes. For more details see \cite{Petrovic2021}.
  
  \section{Additional results}
  \label{sec:appendix:additional}
  
  In the following, we provide additional experimental results, namely the optimal order of a $k$-th order De Bruijn graph model, inferred using the statistical model selection approach from \cite{Scholtes2017} (\cref{tab:order}), additional results for the number of hits among the top-ranked nodes for betweenness and closeness centrality (\cref{tab:appendix:hits:betweenness} and \cref{tab:appendix:hits:closeness}). We also provide the MAE scores in tables \ref{tab:appendix:MAE:bw} and \ref{tab:appendix:MAE:cl} for the betweenness and closeness centrality across all models.
  
  \begin{table}[htbp]
    
    \caption{Result of detection of optimal order based on likelihood ratio test.}
    \label{tab:order}
      \centering
      \begin{tabular}{l|c c}%
        \toprule 
      data set &  $K_{opt}$ train&  $K_{opt}$ val \\ 
      \midrule 
      ants-1-1 & 2 & 2 \\ 
      ants-1-2 & 2 & 2 \\ 
      ants-2-1 & 1 & 1 \\ 
      ants-2-2 & 2 & 2 \\ 
      company-emails & 2 & 2 \\ 
      eu-email-4 & 1 & 1 \\ 
      eu-email-2 & 2 & 2 \\ 
      eu-email-3 & 1 & 1 \\ 
      sp-hospital & 2 & 2 \\ 
      sp-hypertext & 2 & 2 \\ 
      sp-workplace & 2 & 2 \\ 
      sp-highschool & 2 & 2 \\ 
      haggle & 2 & 2 \\ 
      \bottomrule 
      \end{tabular}
    \end{table}

    \begin{table*}[!htbp]
      \caption{Results for hitsIn5 and hitsIn10 for prediction of temporal betweenness centrality and learning rate for which each experiment performed best}
        \label{tab:appendix:hits:betweenness}
        \centering
        \resizebox{\textwidth}{!}{%
        \begin{tabular}{l|rll|rll}
\toprule
  & \multicolumn{3}{|c}{DBGNN} & \multicolumn{3}{|c}{GCN} \\
Experiment & lr & hitsIn30 & hitsIn5 & lr & hitsIn30 & hitsIn5 \\
\midrule
ants-1-1 & 0.001 & 17.55 ± 1.468 & 2.45 ± 0.887 & 0.001 & 14 ± 2.34 & 0.75 ± 0.786 \\
ants-1-2 & 0.001 & 20.7 ± 1.867 & 2.1 ± 0.553 & 0.001 & 18.6 ± 1.984 & 1.7 ± 0.801 \\
ants-2-1 & 0.001 & 17.3 ± 1.342 & 1.85 ± 0.813 & 0.001 & 14.6 ± 2.137 & 0.8 ± 0.834 \\
ants-2-2 & 0.100 & 18.3 ± 1.867 & 1.05 ± 0.605 & 0.001 & 14.8 ± 3.915 & 0.55 ± 0.686 \\
eu-email-dept4 & 0.010 & 14.2 ± 3.205 & 1.2 ± 1.152 & 0.100 & 9.8 ± 5.217 & 0.3 ± 0.733 \\
eu-email-dept2 & 0.010 & 14.65 ± 1.785 & 1.4 ± 1.046 & 0.010 & 15.1 ± 2.315 & 1.3 ± 0.801 \\
eu-email-dept3 & 0.010 & 18.1 ± 5.418 & 2.75 ± 1.209 & 0.100 & 17.75 ± 1.713 & 3.1 ± 0.912 \\
sp-workplace & 0.100 & 20.1 ± 1.586 & 1.85 ± 0.489 & 0.001 & 16.25 ± 1.517 & 1.1 ± 0.852 \\
sp-hypertext & 0.100 & 21.6 ± 1.353 & 2.1 ± 0.718 & 0.100 & 20.6 ± 0.821 & 3.2 ± 0.523 \\
sp-hospital & 0.010 & 24.9 ± 1.071 & 2.45 ± 0.51 & 0.010 & 24.65 ± 0.671 & 2.25 ± 0.55 \\
haggle & 0.001 & 25.75 ± 1.333 & 1.1 ± 0.852 & 0.001 & 26 ± 0.0 & 0 ± 0.0 \\
manufacturing-email & 0.010 & 19.05 ± 2.188 & 1.05 ± 0.826 & 0.100 & 13.7 ± 2.342 & 0.15 ± 0.489 \\
sp-highschool-2013 & 0.100 & 13.7 ± 1.922 & 1.3 ± 0.865 & 0.001 & 7.2 ± 2.118 & 0.25 ± 0.444 \\
\bottomrule
\end{tabular}

        }
      \end{table*}

    \begin{table*}[!htbp]
      \caption{Results for hitsIn5 and hitsIn10 for prediction of temporal closeness centrality and learning rate for which each experiment performed best}
        \label{tab:appendix:hits:closeness}
        \centering
        \resizebox{\textwidth}{!}{%
        \begin{tabular}{l|rll|rll}
\toprule
  & \multicolumn{3}{|c}{DBGNN} & \multicolumn{3}{|c}{GCN} \\
Experiment & lr & hitsIn30 & hitsIn5 & lr & hitsIn30 & hitsIn5 \\
\midrule
ants-1-1 & 0.100 & 25.7 ± 0.865 & 3.45 ± 0.51 & 0.001 & 25.65 ± 0.587 & 2 ± 0.0 \\
ants-1-2 & 0.010 & 27.2 ± 0.894 & 3.95 ± 0.224 & 0.001 & 19.9 ± 0.308 & 3 ± 0.0 \\
ants-2-1 & 0.100 & 28.05 ± 0.394 & 5 ± 0.0 & 0.001 & 24 ± 0.0 & 4 ± 0.0 \\
ants-2-2 & 0.001 & 26.35 ± 0.671 & 3.6 ± 0.681 & 0.010 & 23.1 ± 0.641 & 1.35 ± 0.587 \\
eu-email-dept4 & 0.001 & 24.25 ± 1.164 & 3.9 ± 0.308 & 0.100 & 10.25 ± 4.327 & 0.45 ± 0.759 \\
eu-email-dept2 & 0.100 & 22.45 ± 1.099 & 4.85 ± 0.366 & 0.001 & 11 ± 0.0 & 2 ± 0.0 \\
eu-email-dept3 & 0.010 & 27.65 ± 0.489 & 4 ± 0.0 & 0.001 & 19.85 ± 0.587 & 3 ± 0.0 \\
sp-workplace & 0.001 & 26.35 ± 0.813 & 2.9 ± 0.308 & 0.001 & 16.65 ± 0.489 & 3 ± 0.0 \\
sp-hypertext & 0.001 & 26.3 ± 0.923 & 4.45 ± 0.51 & 0.001 & 18 ± 0.0 & 3.2 ± 0.41 \\
sp-hospital & 0.001 & 24.25 ± 0.444 & 3.1 ± 0.553 & 0.001 & 24 ± 0.0 & 2 ± 0.0 \\
haggle & 0.100 & 28.9 ± 0.447 & 4.3 ± 0.47 & 0.001 & 28 ± 0.0 & 0.45 ± 0.759 \\
manufacturing-email & 0.100 & 26.25 ± 0.639 & 4.55 ± 0.51 & 0.001 & 19 ± 0.0 & 0.5 ± 0.513 \\
sp-highschool-2013 & 0.001 & 22.45 ± 0.686 & 3.7 ± 0.47 & 0.001 & 9.95 ± 0.224 & 0 ± 0.0 \\
\bottomrule
\end{tabular}

        }
    \end{table*}

    \begin{table*}[!htbp]
      \caption{MAE scores for betweenness centralities}
        \label{tab:appendix:MAE:bw}
        \centering
        \resizebox{\textwidth}{!}{%
        \begin{tabular}{l|lllll}
\toprule
   & DBGNN & GCN & EVO & TGN \\
\midrule
 ants-1-1 & 237.617 ± 0.512 & 246.328 ± 0.524 & 122.885 ± 14.865 & 61.646 ± 1.978 \\
 ants-1-2 & 23.272 ± 1.344 & 33.6 ± 2.015 & 145.597 ± 0.663 & 31.422 ± 2.085 \\
 ants-2-1 & 3.346 ± 0.229 & 5.395 ± 1.01 & 147.795 ± 23.11 & 5.243 ± 0.188 \\
 ants-2-2 & 34.481 ± 1.55 & 36.855 ± 0.217 & 76.492 ± 28.231 & 30.475 ± 6.468 \\
 eu-email-dept4 & 5.512 ± 0.372 & 9.345 ± 3.485 & 174.706 ± 2.694 & 3.044 ± 0.167 \\
 eu-email-dept2 & 3.048 ± 0.313 & 4.824 ± 2.114 & 132.947 ± 4.404 & 2.097 ± 0.08 \\
 eu-email-dept3 & 2.285 ± 0.108 & 3.019 ± 0.478 & 209.543 ± 1.928 & 2.248 ± 0.249 \\
 sp-workplace & 70.803 ± 3.242 & 83.534 ± 0.18 & 89.494 ± 9.508 & 2.334 ± 0.236 \\
 sp-hypertext & 66.583 ± 3.113 & 137.715 ± 0.332 & 257.435 ± 12.054 & 98.36 ± 24.553 \\
 sp-hospital & 24.254 ± 1.505 & 43.885 ± 0.188 & 87.532 ± 8.734 & 6.864 ± 0.667 \\
 haggle & 11.282 ± 0.313 & 14.85 ± 0.597 & 227.847 ± 4.947 & 12.542 ± 0.673 \\
 manufacturing-email & 59.722 ± 4.245 & 74.79 ± 0.128 & 180.904 ± 54.201 & 38.472 ± 1.825 \\
 sp-highschool-2013 & 496.086 ± 15.04 & 813.573 ± 0.338 & 781.268 ± 11.873 & 2.754 ± 0.028 \\
\bottomrule
\end{tabular}

        }
    \end{table*}

    \begin{table*}[!htbp]
      \caption{MAE scores for closeness centralities}
        \label{tab:appendix:MAE:cl}
        \centering
        \resizebox{\textwidth}{!}{%
        \begin{tabular}{l|lllll}
\toprule
 &    DBGNN & GCN & EVO & TGN \\
\midrule
ants-1-1 & 1164.711 ± 14.402 & 1597.068 ± 0.392 & 122.885 ± 14.865 & 372.877 ± 18.343 \\
ants-1-2 & 183.19 ± 7.212 & 870.068 ± 0.33 & 145.597 ± 0.663 & 262.986 ± 16.023 \\
 ants-2-1 & 50.552 ± 2.196 & 406.514 ± 0.379 & 147.795 ± 23.11 & 173.593 ± 2.885 \\
 ants-2-2 & 120.622 ± 7.597 & 581.628 ± 0.596 & 76.492 ± 28.231 & 264.237 ± 17.492 \\
 eu-email-dept4 & 152.888 ± 10.107 & 603.253 ± 86.201 & 174.706 ± 2.694 & 254.053 ± 13.355 \\
 eu-email-dept2 & 277.821 ± 28.434 & 1285.315 ± 0.327 & 132.947 ± 4.404 & 400.356 ± 42.539 \\
 eu-email-dept3 & 70.476 ± 5.577 & 807.134 ± 0.285 & 209.543 ± 1.928 & 320.185 ± 4.476 \\
 sp-workplace & 343.833 ± 12.627 & 1646.676 ± 0.377 & 89.494 ± 9.508 & 297.783 ± 30.86 \\
 sp-hypertext & 1481.244 ± 48.473 & 5804.967 ± 0.312 & 257.435 ± 12.054 & 1559.56 ± 146.031 \\
 sp-hospital & 600.69 ± 17.705 & 2286.325 ± 0.438 & 87.532 ± 8.734 & 244.128 ± 8.619 \\
 haggle & 365.089 ± 17.982 & 2404.175 ± 0.412 & 227.847 ± 4.947 & 1507.635 ± 69.119 \\
 manufacturing-email & 1244.012 ± 52.211 & 5652.538 ± 0.219 & 180.904 ± 54.201 & 892.742 ± 13.726 \\
sp-highschool-2013 & 8000.875 ± 514.726 & 32138.809 ± 0.305 & 781.268 ± 11.873 & 417.952 ± 4.803 \\
\bottomrule
\end{tabular}

        }
    \end{table*}

    \section{Details on Hyperparameters and Computational Resources}
    \label{sec:appendix:architecture}
  
  In \cref{tab:architecture} - \cref{tab:architecture3} we provide further details on the neural network architecture that we used for our experiments with DBGNN, GCN and EVO.
  For DBGNN and GCN, we tested different learning rates between $0.1$ and $0.001$ using an ADAM optimizer with a weight decay of $5 \cdot 10^{-4}$.

  For the feed forward neural network applied to the 16-dimensional embeddings generated by EVO, we used a fully connected network with ReLu activation functions, and input layer with 16 dimensions and a hidden layer with eight dimensions. We trained the model for $2000$ epochs testing different learning rates between $0.01$ and $0.0001$ with an ADAM optimizer with weight decay of $5 \cdot 10^{-4}$.

  As hyperparameters of the TGN architecture, we used learning rates between $0.01$ and $0.0001$ with an ADAM optimizer with weight decay of $5 \cdot 10^{-4}$ and 200 epochs. We further tested window sizes $k \in \{3,5,7,10\}$, batch sizes between $100$ and $600$ and memory dimensions between $50$ and $70$.

  All experiments were run on two dedicated workstation machines. The first workstation had an AMD Ryzen 9 7950X 16-core CPU with 32 GB of RAM and Nvidia RTX 4090 GPU. The second machine was equipped with an AMD Ryzen 9 7900X 12-Core CPU with 32 GB of RAM and Nvidia RTX 4080 GPU.
  
    \begin{table*}[!htbp]
      \centering \small
      \begin{tabular}{l c c c }  
      Layer & Input dimensions & Output dimensions & Activation Function\\  
      \toprule      
      GCNConv  & $|V|$ & 16 & Sigmoid \\ 
      GCNConv &16  & 8 & ELU  \\ 
      Linear layer& 8 &  1& ELU \\
      \bottomrule
      \end{tabular}%
      \caption{Overview of proposed model architecture for simple GCN}
      \label{tab:architecture} 
      \end{table*}
    
    \begin{table*}[!htbp]
      \centering \small
      \begin{tabular}{l c c c }  
      Layer & Input dimensions & Output dimensions & Activation Function \\  
      \toprule      
      GCNConv first order  & $|V|$ & 16 & Sigmoid\\ 
      GCNConv second order  & $|E|$ & 16 & Sigmoid\\ 
      Bipartite layer &16 & 8&  ELU\\
      Linear layer & 8&1& ELU\\
      \bottomrule
      \end{tabular}%
      \caption{Overview of proposed model architecture for DBGNN}
      \label{tab:architecture2} 
      \end{table*}

      \begin{table*}[!htbp]
        \centering \small
        \begin{tabular}{l c c c }  
        Layer & Input dimensions & Output dimensions & Activation Function \\  
        \toprule      
        Linear Layer  & 16 & 8 & ReLU\\ 
        Linear Layer  & 8 & 1 & ReLU\\ 
        \bottomrule
        \end{tabular}%
        \caption{Overview of proposed model architecture for EVO}
        \label{tab:architecture3} 
        \end{table*}

      \section{Static vs temporal centralities}
      \label{sec:appendix:static}
      
  Let $G=(V,E)$ be a (static) graph, where $V$ is a set of vertices or nodes and $(v,w) \in E$ are potentially directed edges or links from node $v$ to $w$.
  Let us further consider weighted graphs, where we have a function $w: E \rightarrow \mathbb{N}$ that assigns integer weights to edges.
  In a static network $G=(V,E)$, we define a path (or walk) of length $l$ from $v_0$ to $v_l$ as any sequence of nodes $v_0,\dots,v_l$ iff $(v_{i-1},v_i)\in E$ for $ i=1,\dots,l$.
  If every node occurs only once in the sequence, we call the sequence a \emph{simple} path.
  A shortest path between two nodes $v$ and $w$ is a (not necessarily unique) path of legth $l$ such that all other paths from $v$ to $w$ have length $l'\geq l$.
  
  In static networks, shortest paths between pairs of nodes allow us to define \emph{path-based nodes centralities}, which can be used to identify influential nodes.
  Here, we briefly introduce two important path-based centrality measures, namely \emph{betweenness and closeness centrality}.
  For static networks without temporal interactions the betweenness centrality of a node $v$ is calculated as 
  \[c_B (v)= \sum_{s\neq v\neq t \in V}\frac{\sigma_{s,t}(v)}{\sigma_{s,t}}\]
  where $\sigma_{s,t}$  is the number of the shortest paths between nodes $s$ and $t$ and $\sigma_{s,t}(v)$ is the number of such paths that pass through node $v$. In other words the node is considered central if there are many shortest paths that pass through the node.
  The closeness centrality on the other hand is defined as 
  \[c_C (v)= \frac{1}{\sum_{u\in V}d(u,v)}\]
  where $d(u,v)$ describes the distance (length of the shortest path) of node $u$ to node $v$. Thus, in terms of closeness a node is considered more central if the overall distance to all other nodes in the graph is relatively small.
  
\begin{figure*}[htbp]
  
        \begin{subfigure}[l]{.3\textwidth}
        \includegraphics*[width=\textwidth]{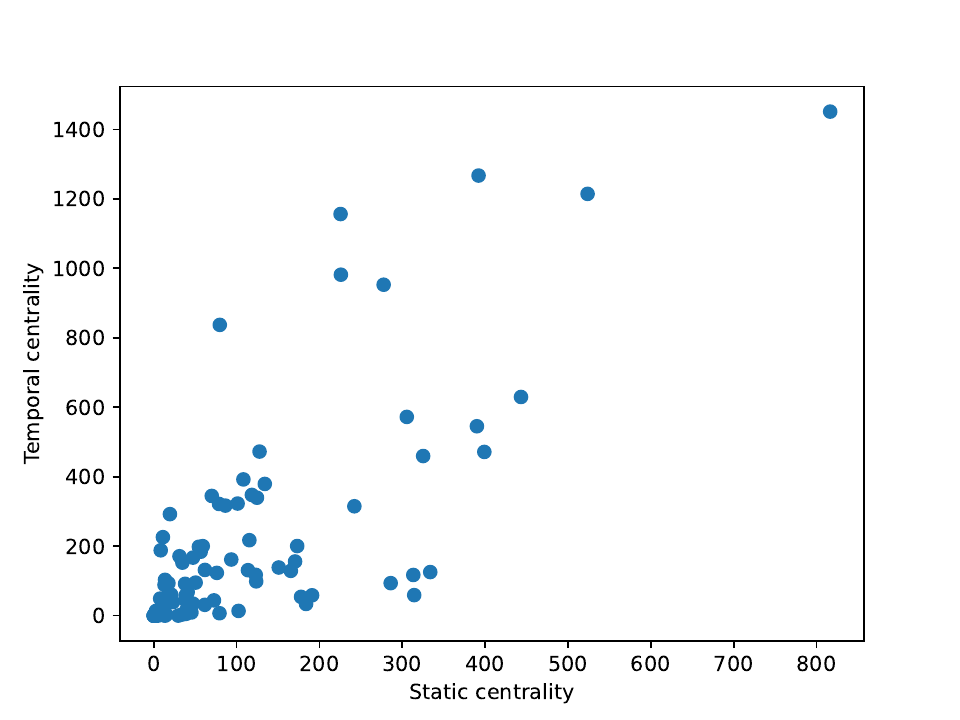}
        \subcaption{ants-1-1}
        \end{subfigure}
        \begin{subfigure}[l]{.3\textwidth}
          \includegraphics*[width=\textwidth]{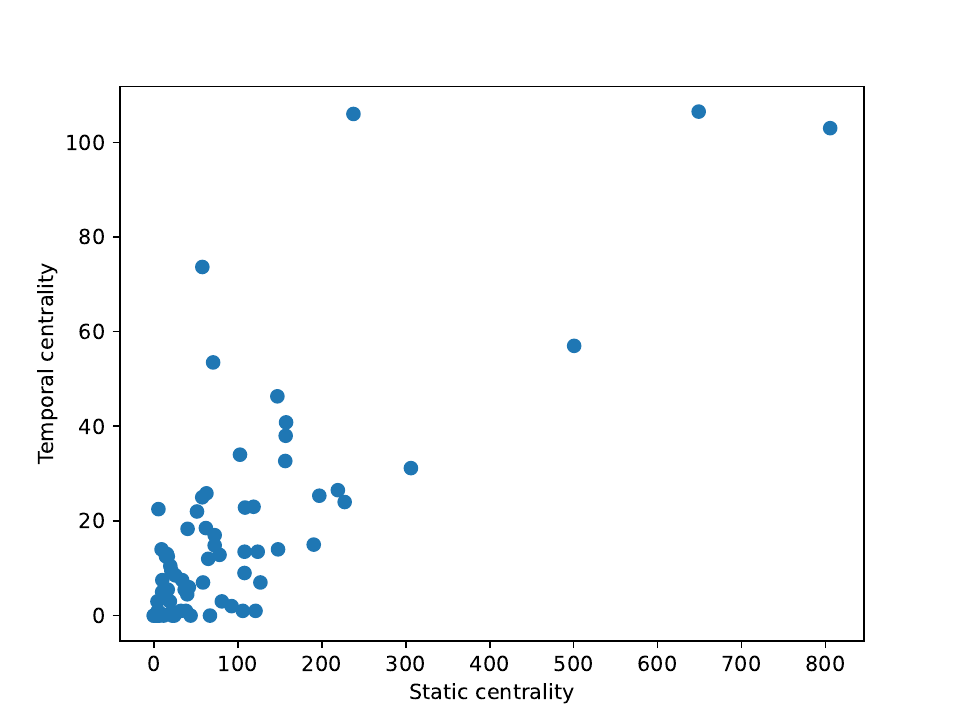}
          \subcaption{ants-1-2}
        \end{subfigure}
        \begin{subfigure}[l]{.3\textwidth}
          \includegraphics*[width=\textwidth]{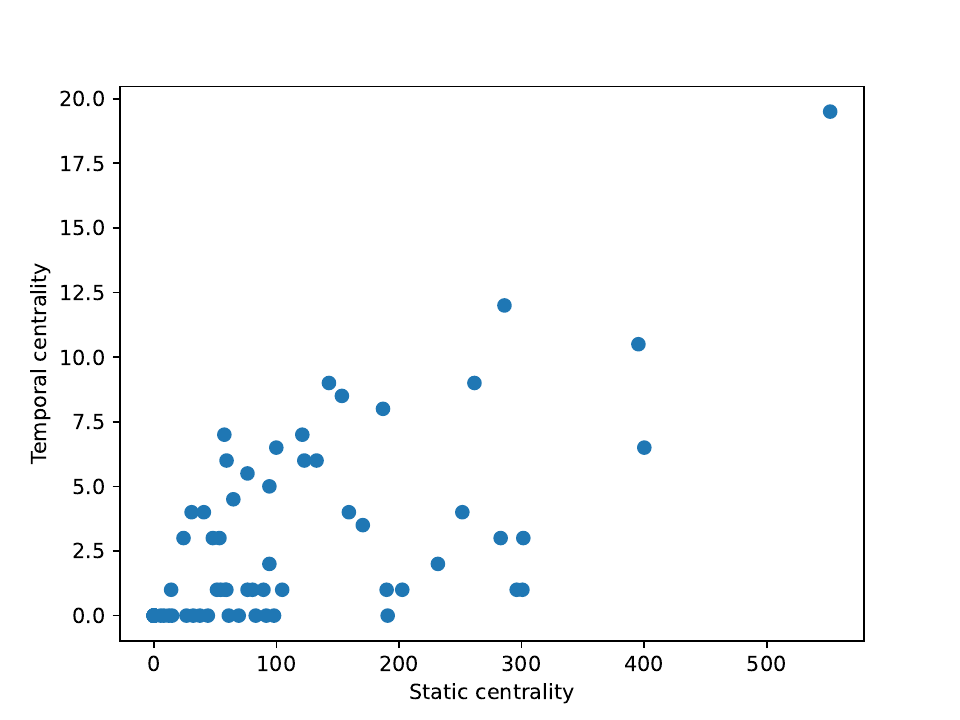}
          \subcaption{ants-2-1}
        \end{subfigure}
  
        \begin{subfigure}[l]{.3\textwidth}
        \includegraphics*[width=\textwidth]{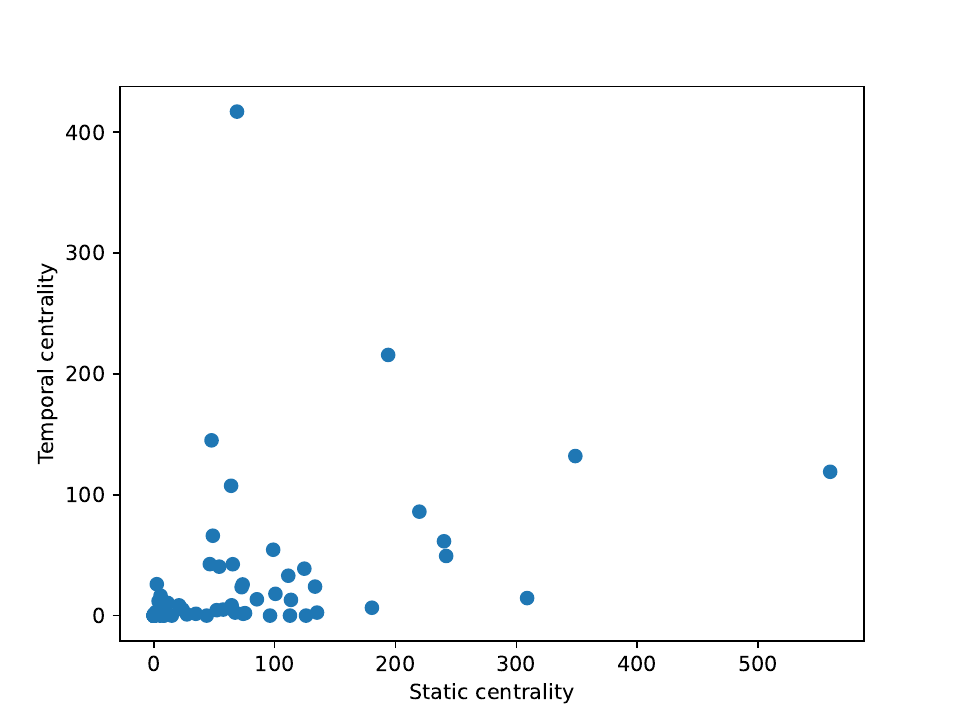}
        \subcaption{ants-2-2}
        \end{subfigure}
        \begin{subfigure}[l]{.3\textwidth}
          \includegraphics*[width=\textwidth]{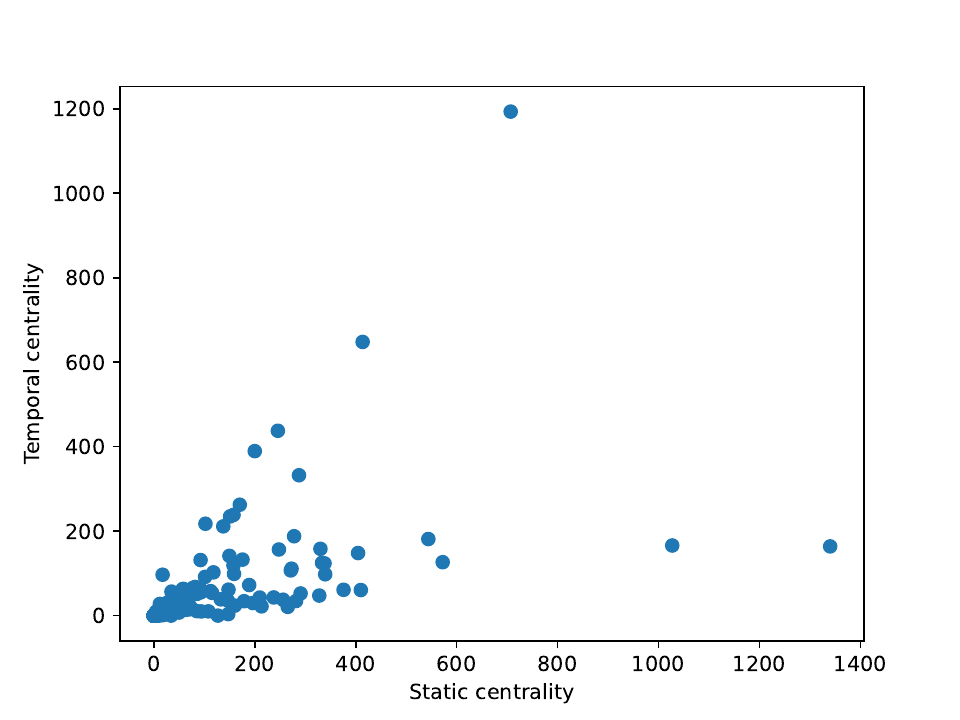}
          \subcaption{company-emails}
        \end{subfigure}
        \begin{subfigure}[l]{.3\textwidth}
          \includegraphics*[width=\textwidth]{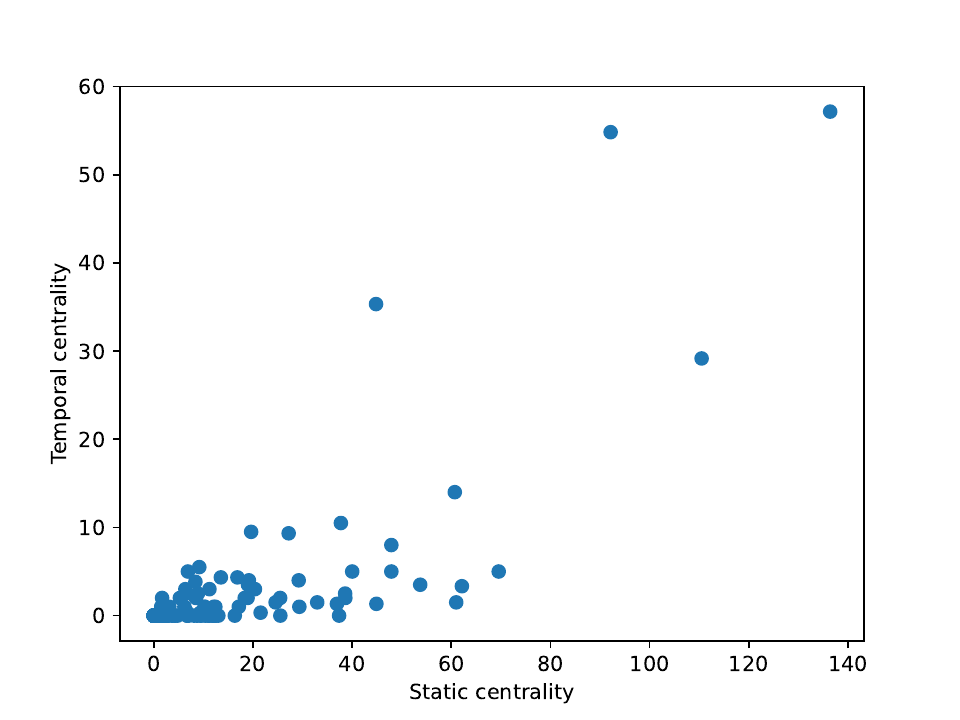}
          \subcaption{eu-email-dept2}
        \end{subfigure}
  
        \begin{subfigure}[l]{.3\textwidth}
          \includegraphics*[width=\textwidth]{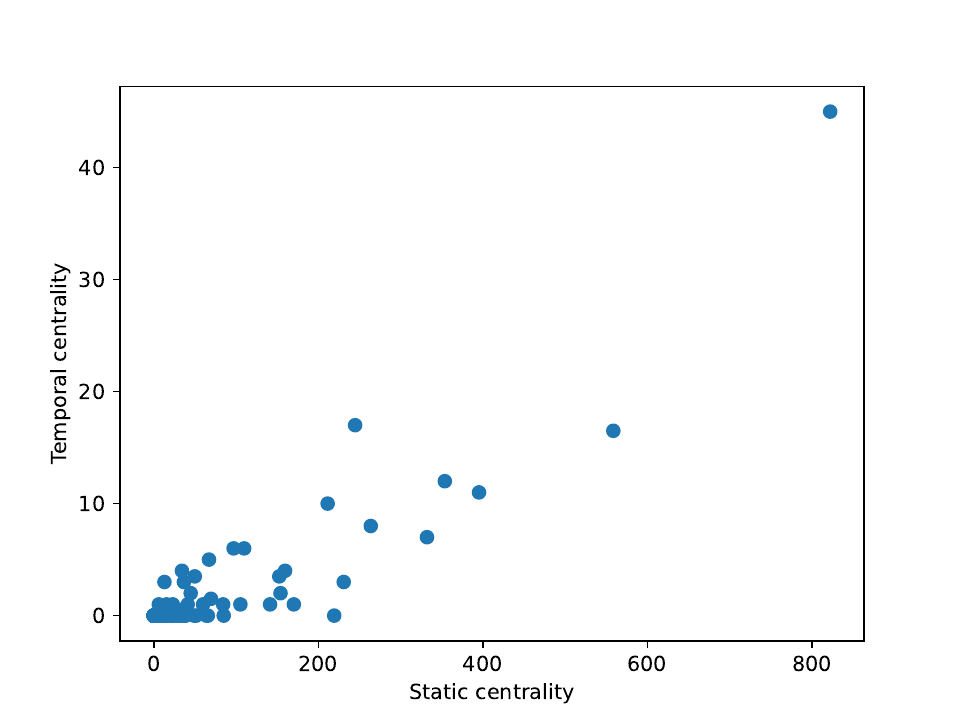}
          \subcaption{eu-email-dept3}
        \end{subfigure}
        \begin{subfigure}[l]{.3\textwidth}
          \includegraphics*[width=\textwidth]{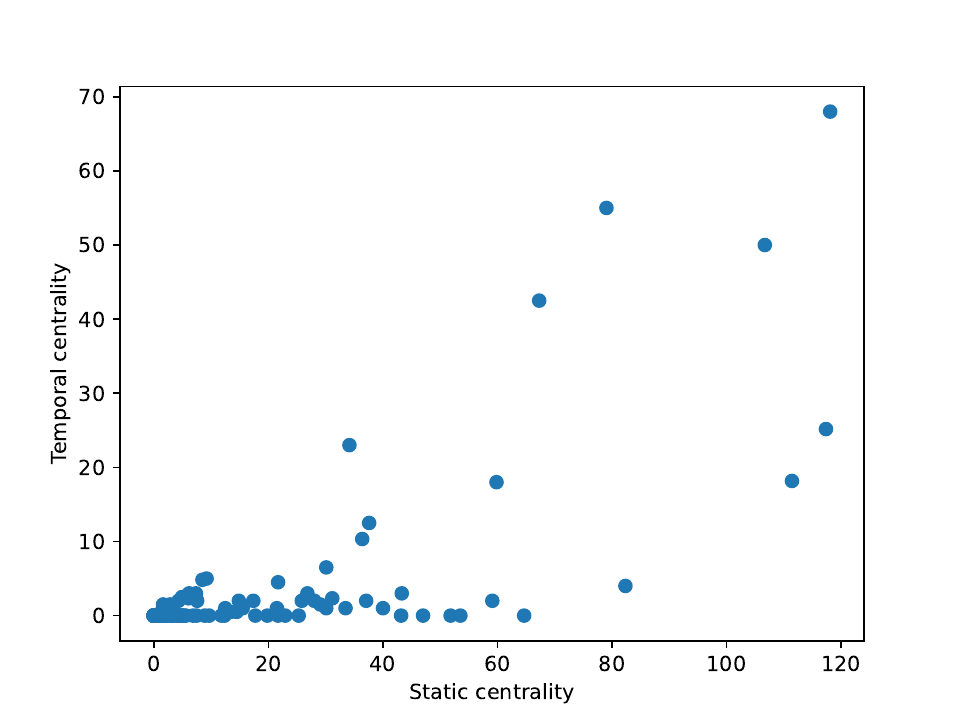}
          \subcaption{eu-email-dept4}
        \end{subfigure}
        \begin{subfigure}[l]{.3\textwidth}
          \includegraphics*[width=\textwidth]{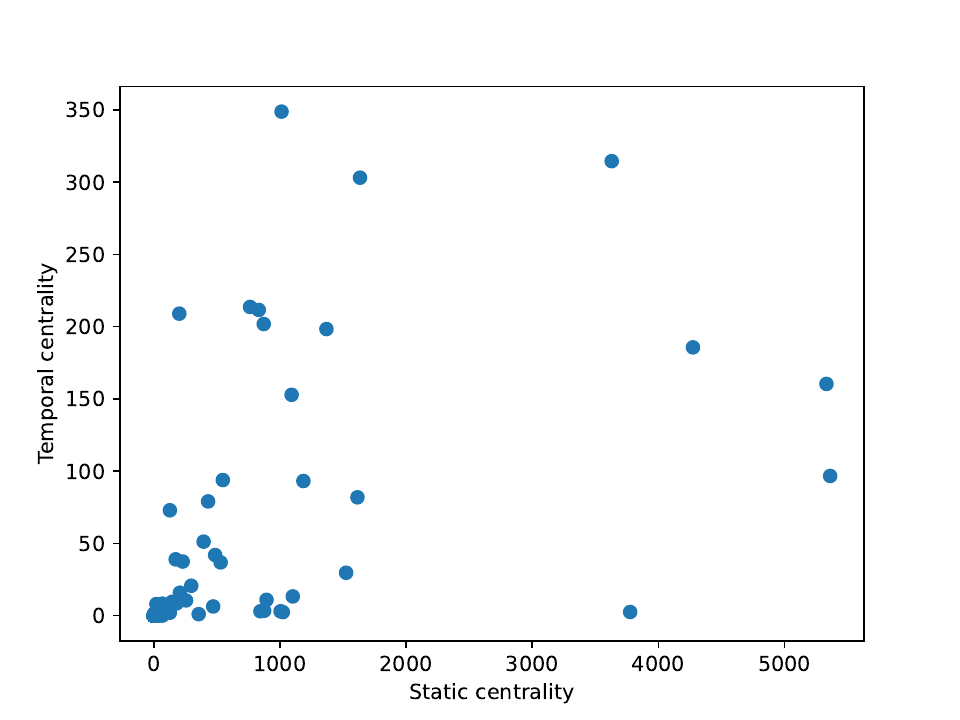}
          \subcaption{haggle}
        \end{subfigure}
  
        \begin{subfigure}[l]{.3\textwidth}
          \includegraphics*[width=\textwidth]{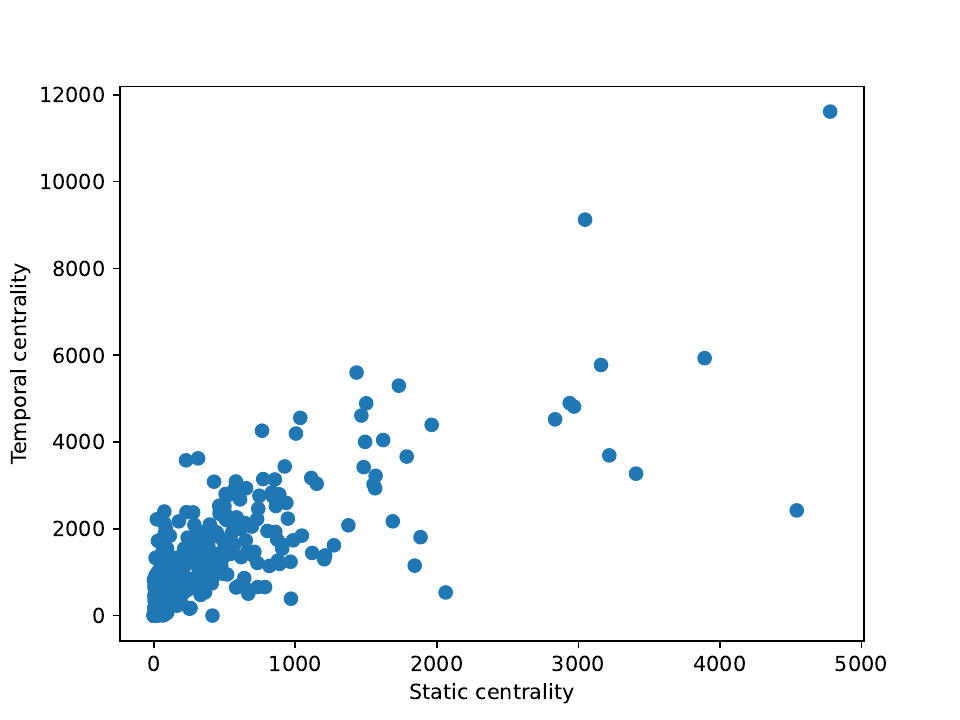}
          \subcaption{sp-highschool}
        \end{subfigure}
        \begin{subfigure}[l]{.3\textwidth}
          \includegraphics*[width=\textwidth]{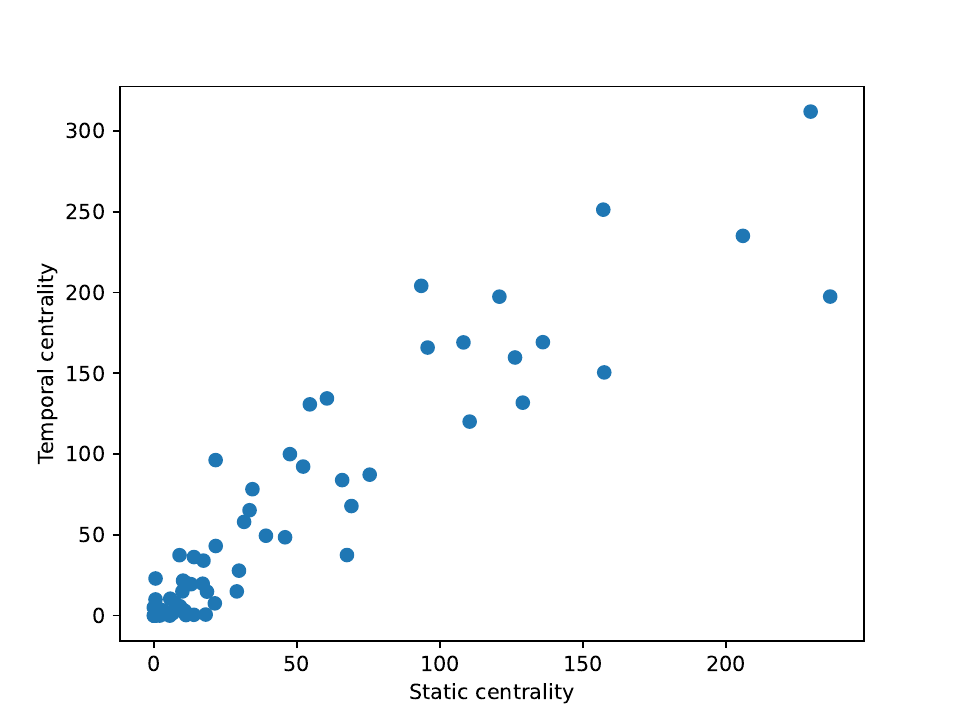}
          \subcaption{sp-hospital}
        \end{subfigure}
        \begin{subfigure}[l]{.3\textwidth}
          \includegraphics*[width=\textwidth]{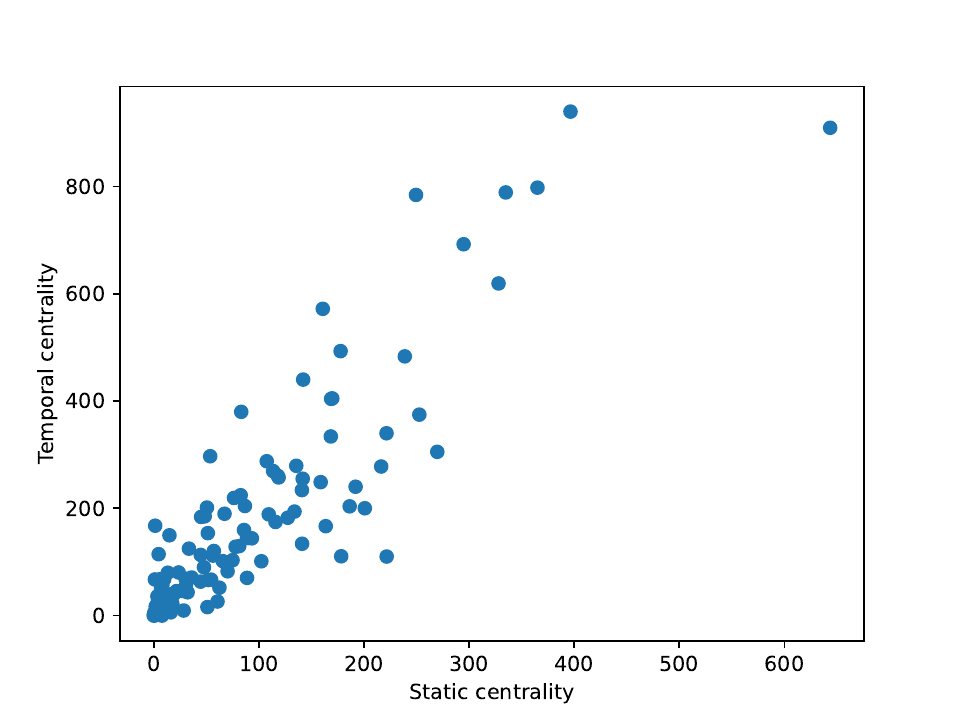}
          \subcaption{sp-hypertext}
        \end{subfigure}
  
        \begin{subfigure}[l]{.3\textwidth}
          \includegraphics*[width=\textwidth]{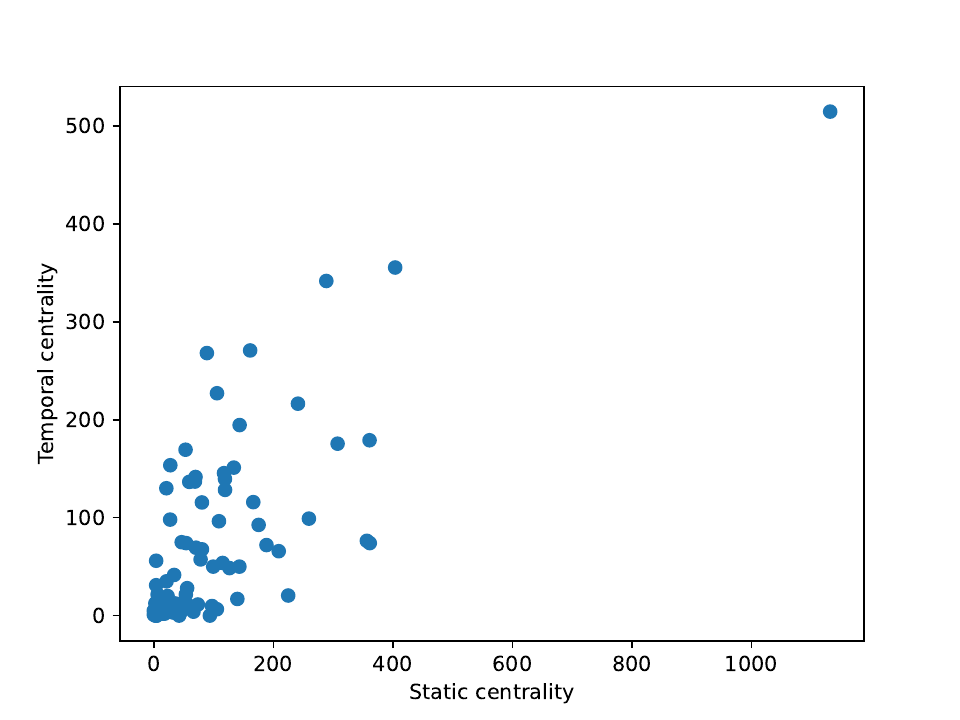}
          \subcaption{sp-workplace}
        \end{subfigure}
  
      \caption{Static vs temporal betweenness centralities of all nodes in 13 empirical dynamic graphs}
      \label{fig:betweenness:corr}
      \end{figure*}

\begin{figure*}[htbp]
  
        \begin{subfigure}[l]{.3\textwidth}
        \includegraphics*[width=\textwidth]{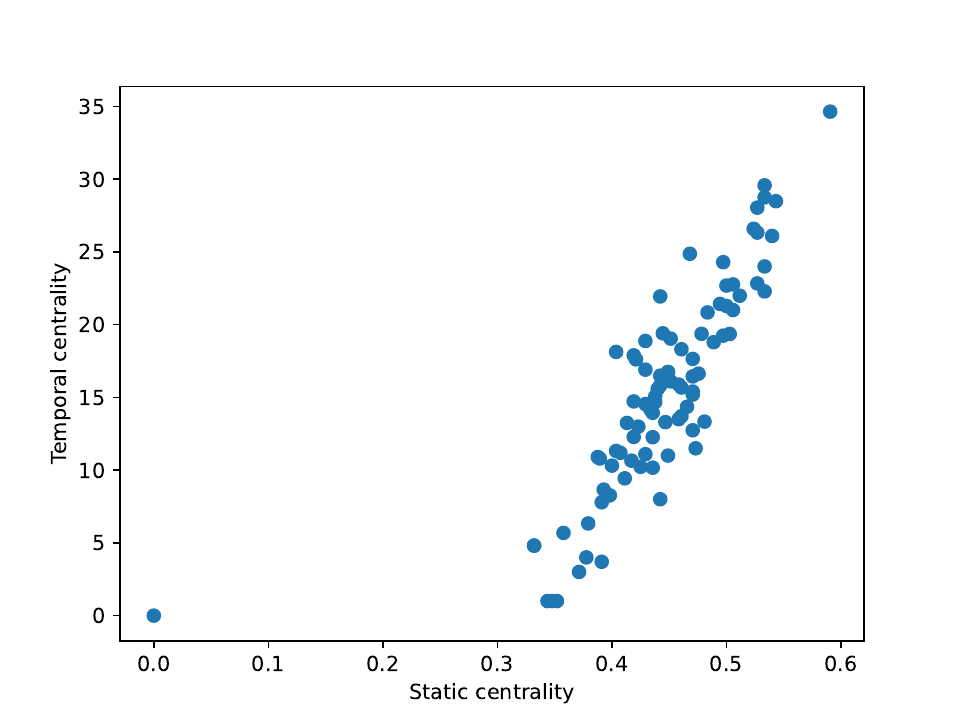}
        \subcaption{ants-1-1}
        \end{subfigure}
        \begin{subfigure}[l]{.3\textwidth}
          \includegraphics*[width=\textwidth]{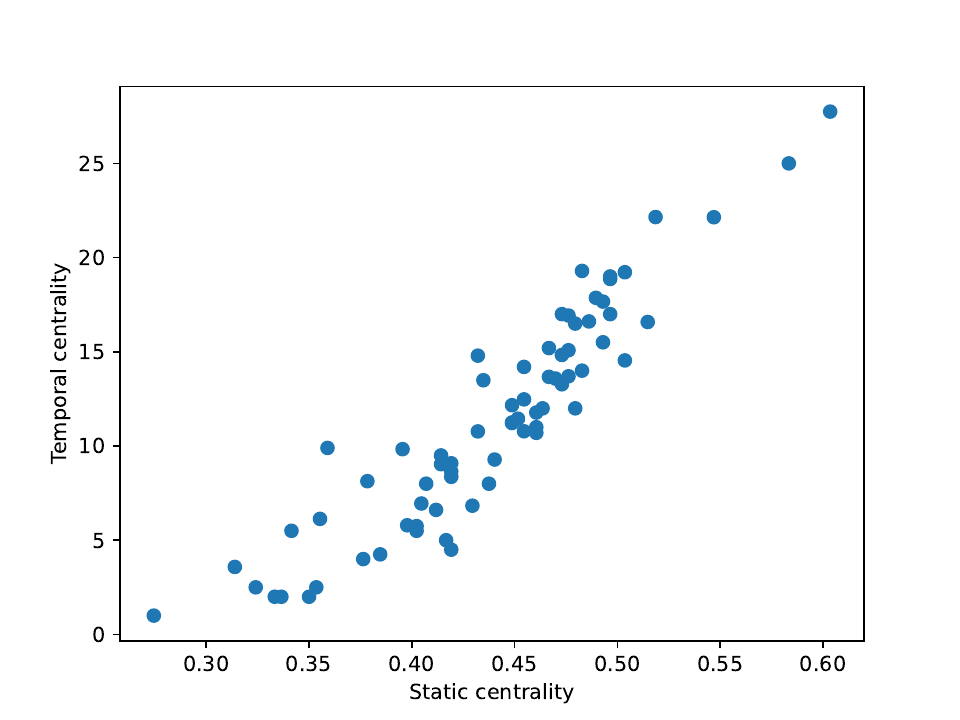}
          \subcaption{ants-1-2}
        \end{subfigure}
        \begin{subfigure}[l]{.3\textwidth}
          \includegraphics*[width=\textwidth]{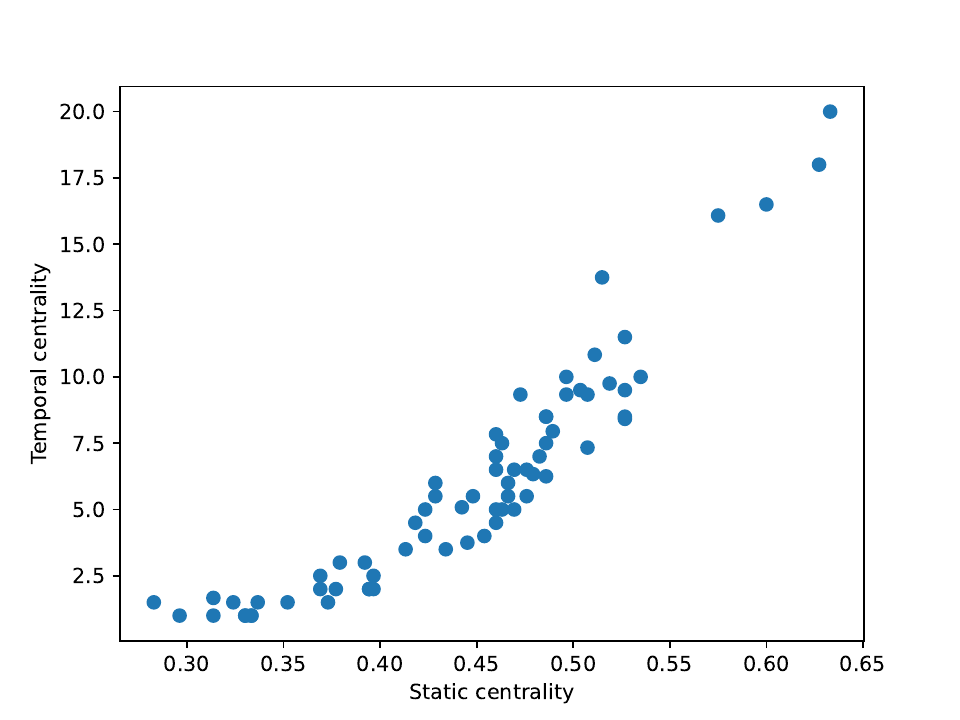}
          \subcaption{ants-2-1}
        \end{subfigure}
  
        \begin{subfigure}[l]{.3\textwidth}
        \includegraphics*[width=\textwidth]{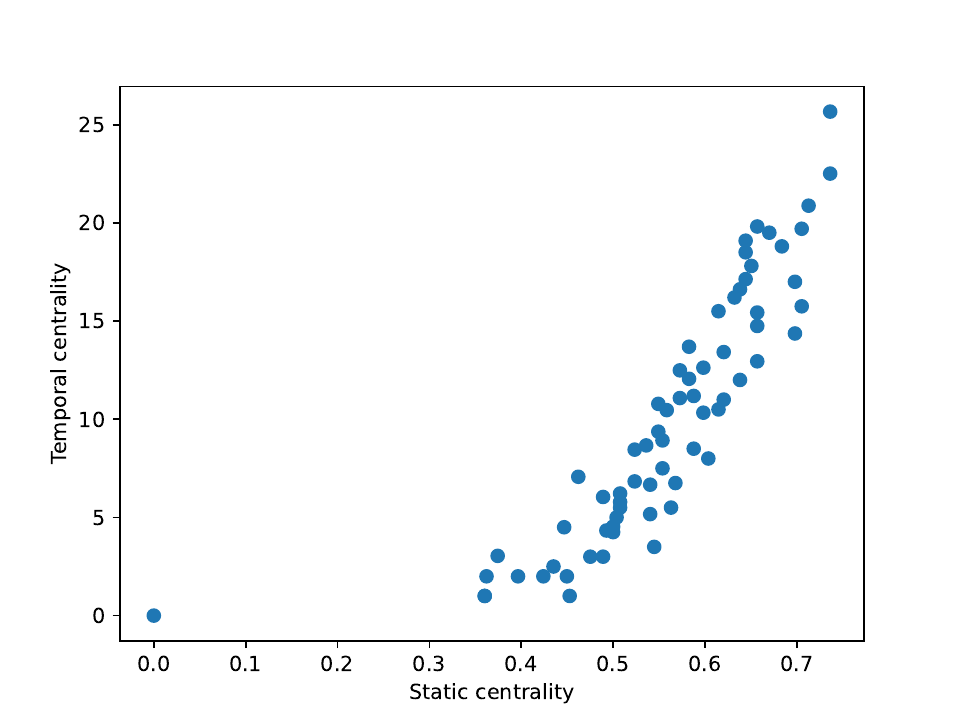}
        \subcaption{ants-2-2}
        \end{subfigure}
        \begin{subfigure}[l]{.3\textwidth}
          \includegraphics*[width=\textwidth]{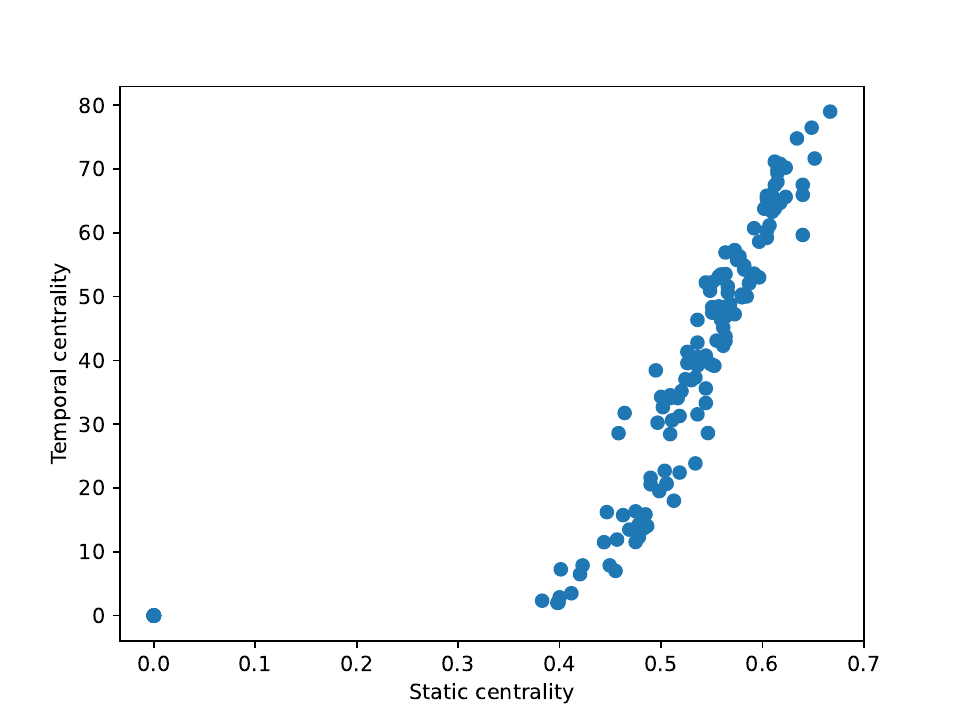}
          \subcaption{company-emails}
        \end{subfigure}
        \begin{subfigure}[l]{.3\textwidth}
          \includegraphics*[width=\textwidth]{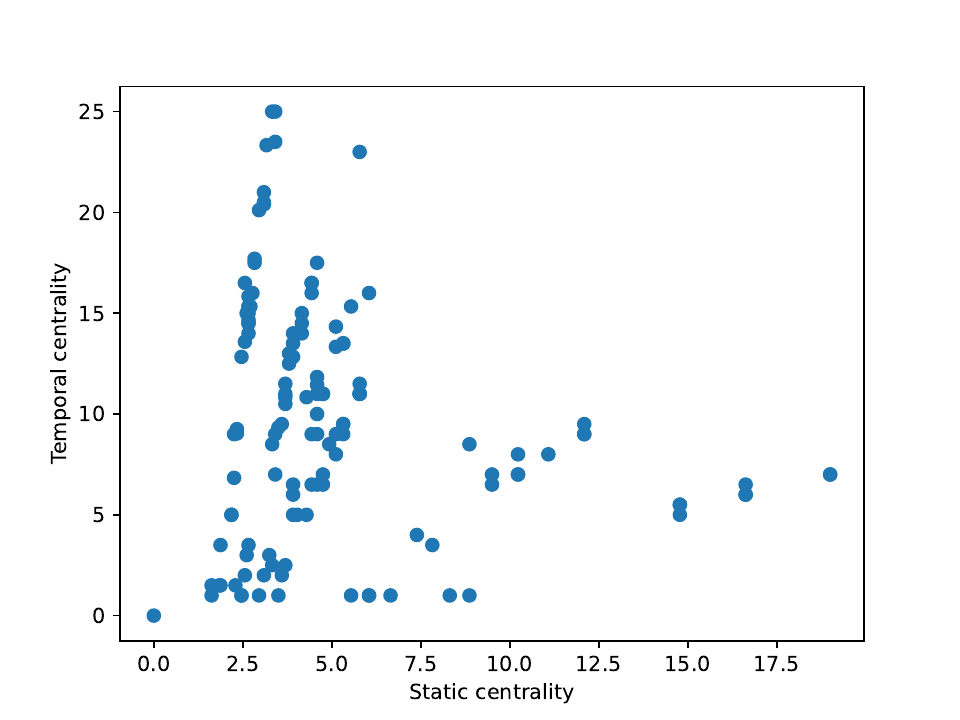}
          \subcaption{eu-email-dept2}
        \end{subfigure}
  
        \begin{subfigure}[l]{.3\textwidth}
          \includegraphics*[width=\textwidth]{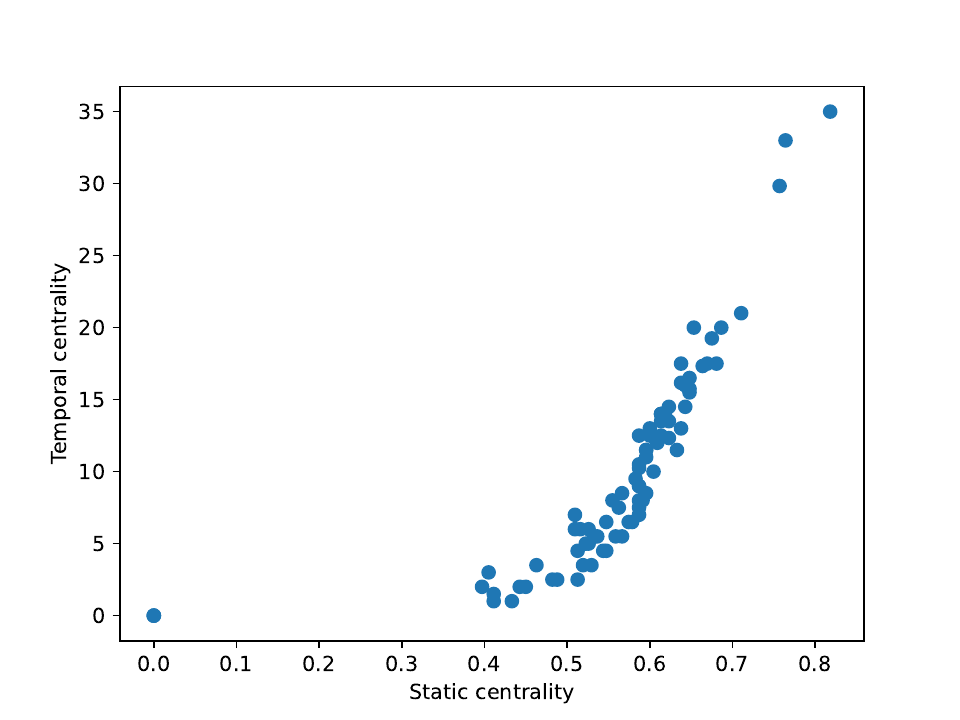}
          \subcaption{eu-email-dept3}
        \end{subfigure}
        \begin{subfigure}[l]{.3\textwidth}
          \includegraphics*[width=\textwidth]{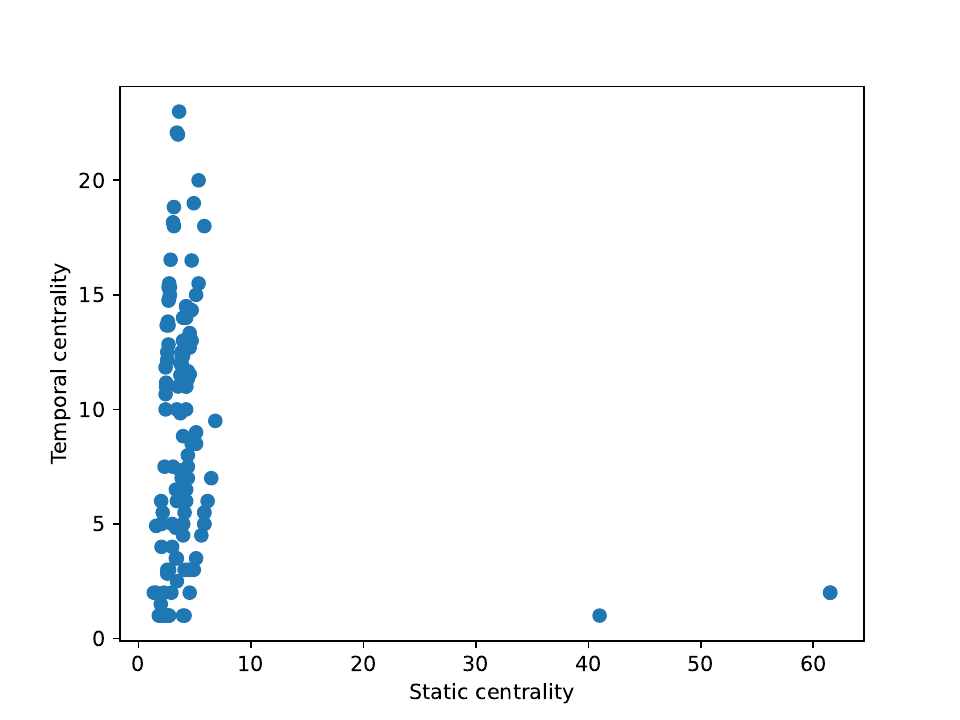}
          \subcaption{eu-email-dept4}
        \end{subfigure}
        \begin{subfigure}[l]{.3\textwidth}
          \includegraphics*[width=\textwidth]{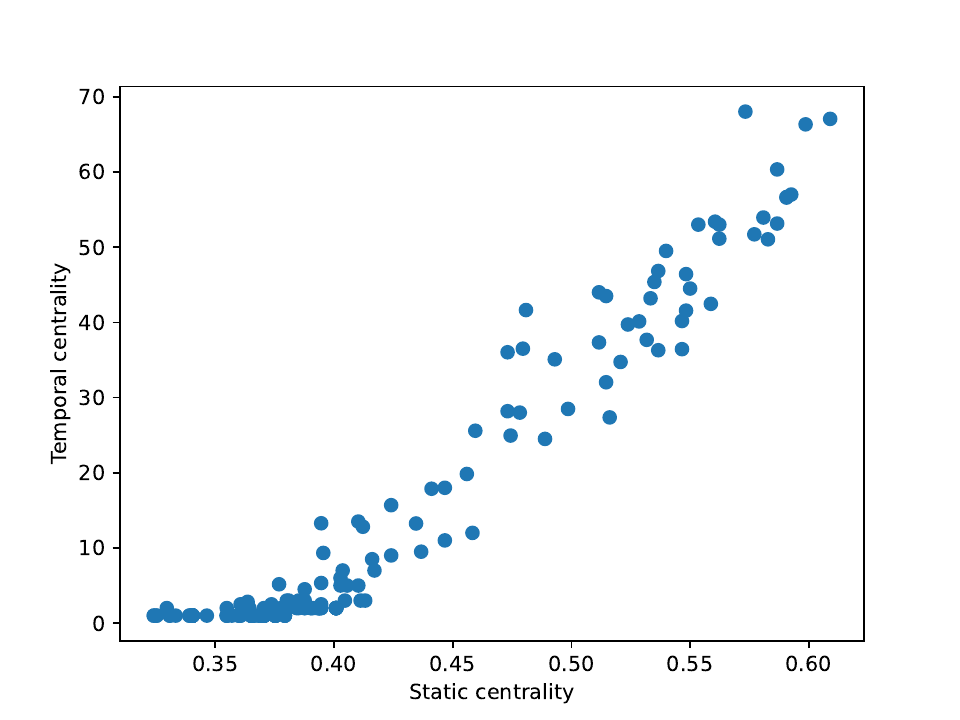}
          \subcaption{haggle}
        \end{subfigure}
  
        \begin{subfigure}[l]{.3\textwidth}
          \includegraphics*[width=\textwidth]{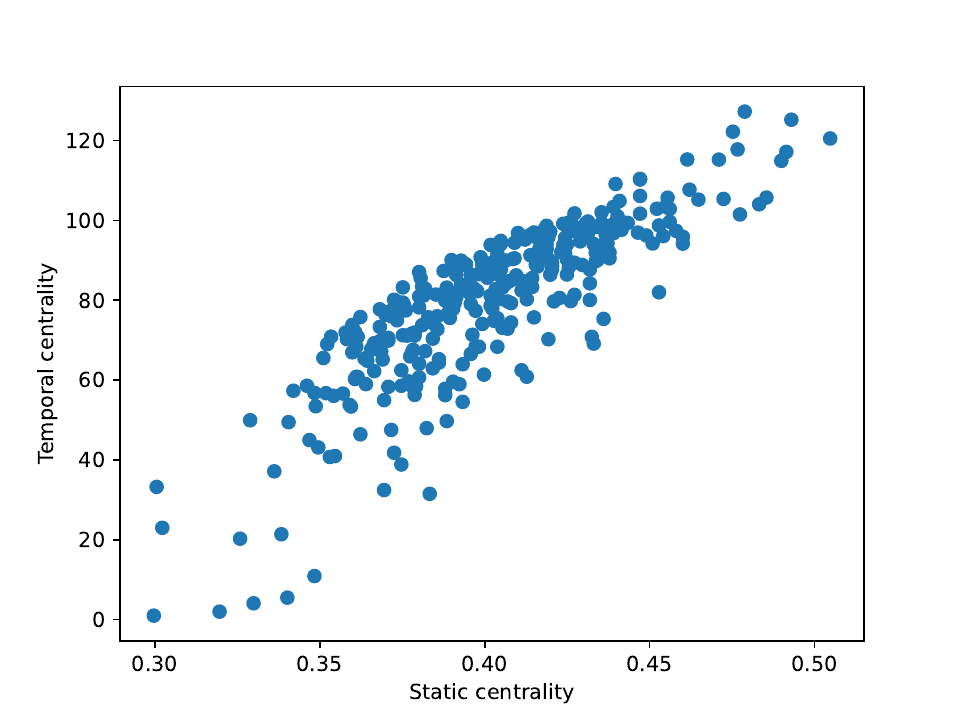}
          \subcaption{sp-highschool}
        \end{subfigure}
        \begin{subfigure}[l]{.3\textwidth}
          \includegraphics*[width=\textwidth]{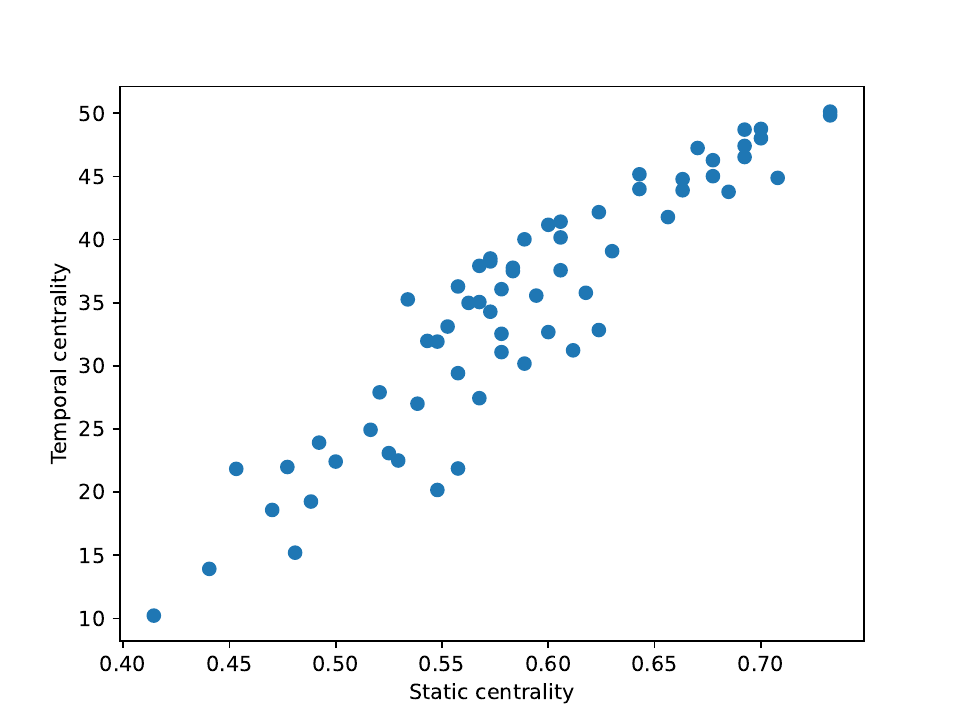}
          \subcaption{sp-hospital}
        \end{subfigure}
        \begin{subfigure}[l]{.3\textwidth}
          \includegraphics*[width=\textwidth]{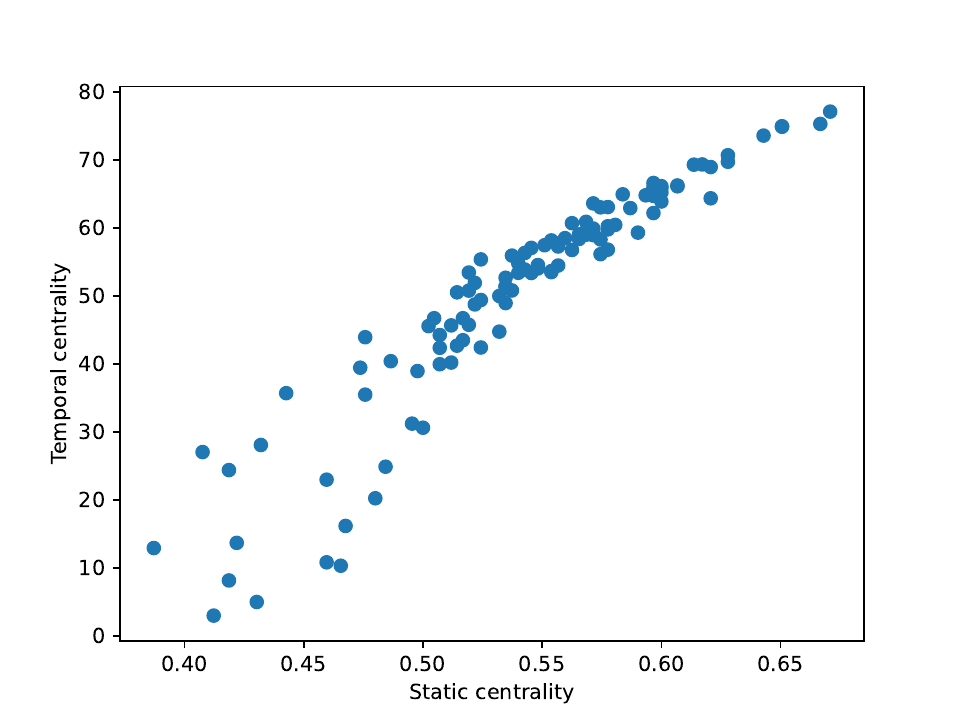}
          \subcaption{sp-hypertext}
        \end{subfigure}
  
        \begin{subfigure}[l]{.3\textwidth}
          \includegraphics*[width=\textwidth]{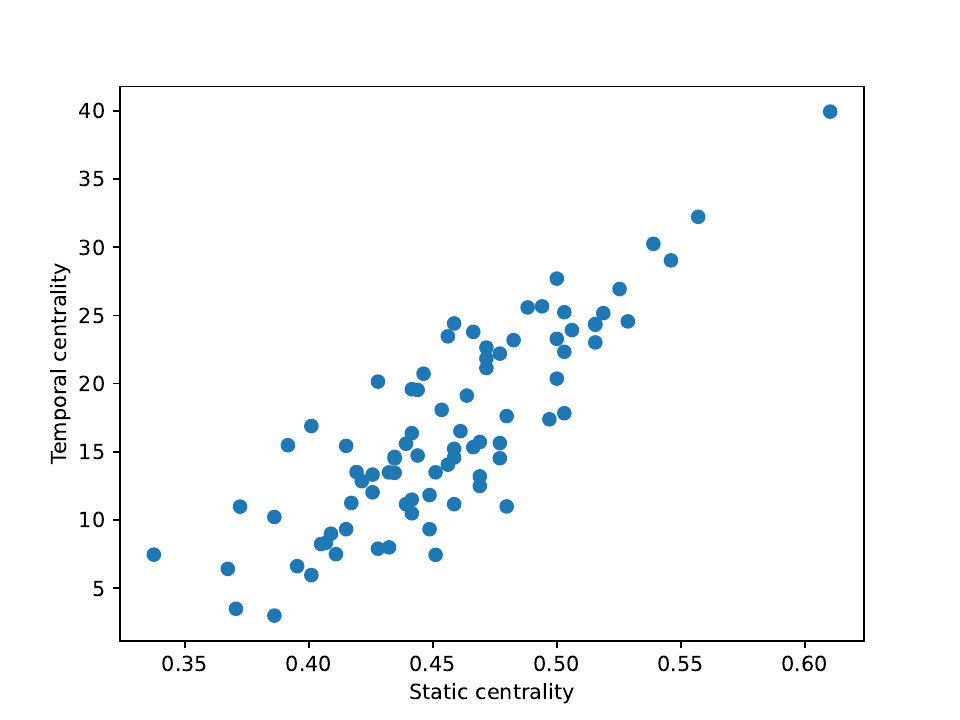}
          \subcaption{sp-workplace}
        \end{subfigure}
  
      \caption{Static vs temporal closeness centralities of all nodes in 13 empirical dynamic graphs}
      \label{fig:closeness:corr}
      \end{figure*}

  To contrast the temporal path-based centralities defined in \cref{sec:relatedwork} with the corresponding static centralities defined above, in \cref{fig:betweenness:corr} and \cref{fig:closeness:corr} we plot the temporal vs. static betweenness and closeness centralities of all nodes for all 13 empirical temporal graphs considered in our work (cf. \cref{tab:data}).
  
      \section{Visualization of Node embeddings}
      \label{sec:embeddings}
  
\begin{figure*}[!htb]
        \begin{subfigure}[c]{.45\textwidth}
        \includegraphics*[width=\textwidth]{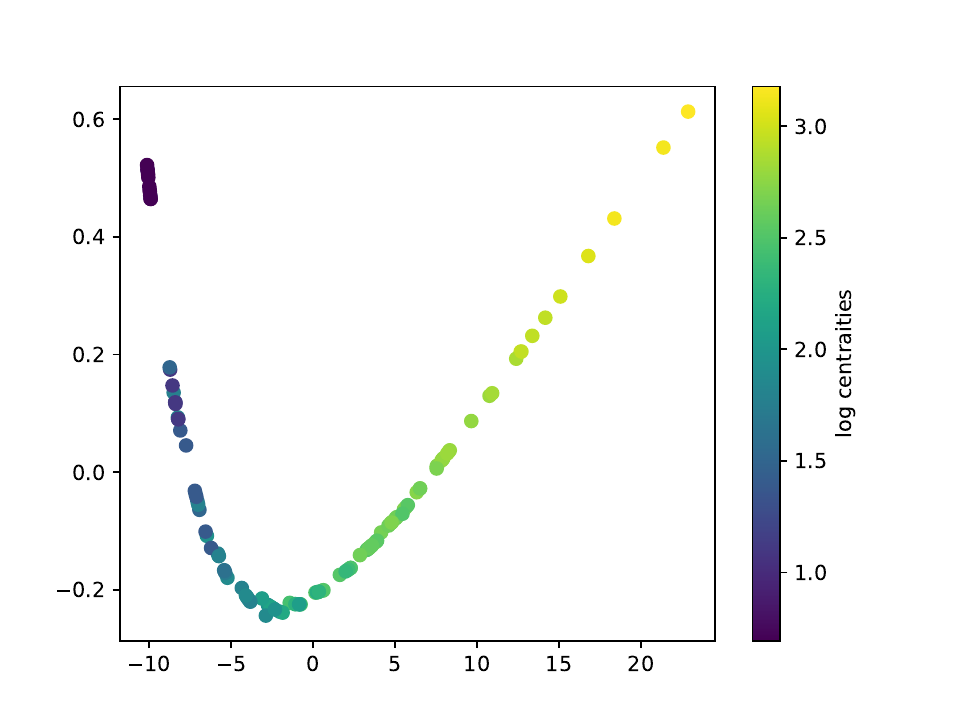}
        \subcaption{Embedding of nodes based on DBGNN model trained for prediction of temporal closeness centrality in eu-email-dept4}
        \end{subfigure}
        \qquad 
        \begin{subfigure}[c]{.45\textwidth}
          \includegraphics*[width=\textwidth]{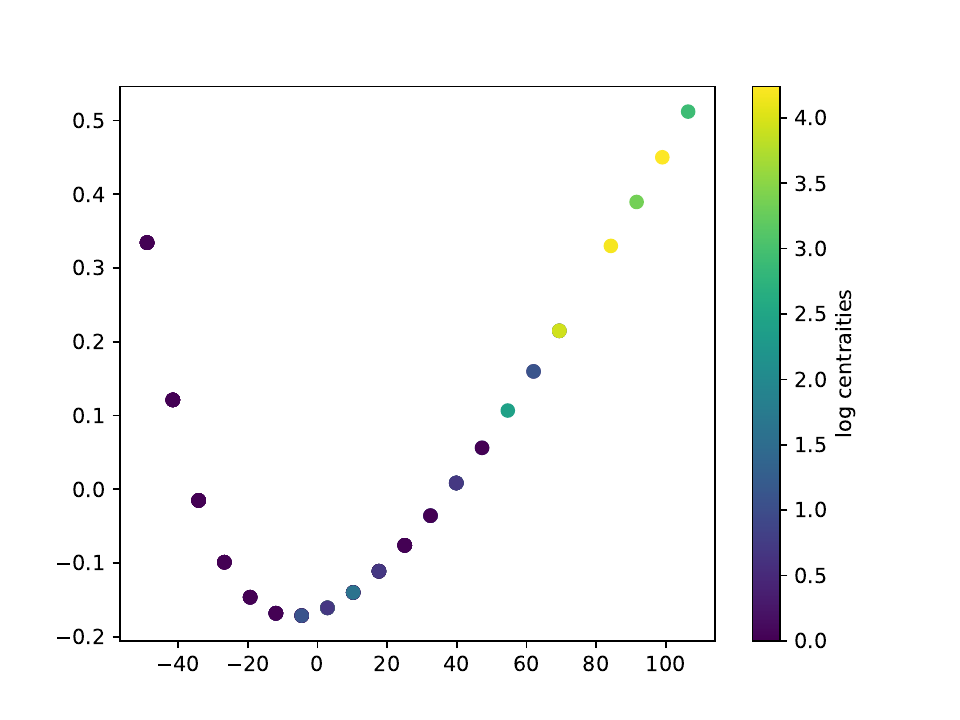}
          \subcaption{Embedding of nodes based on DBGNN model trained for prediction of temporal betweenness centrality in eu-email-dept4}
          \end{subfigure}
  
        \begin{subfigure}[c]{.45\textwidth}
          \includegraphics*[width=\textwidth]{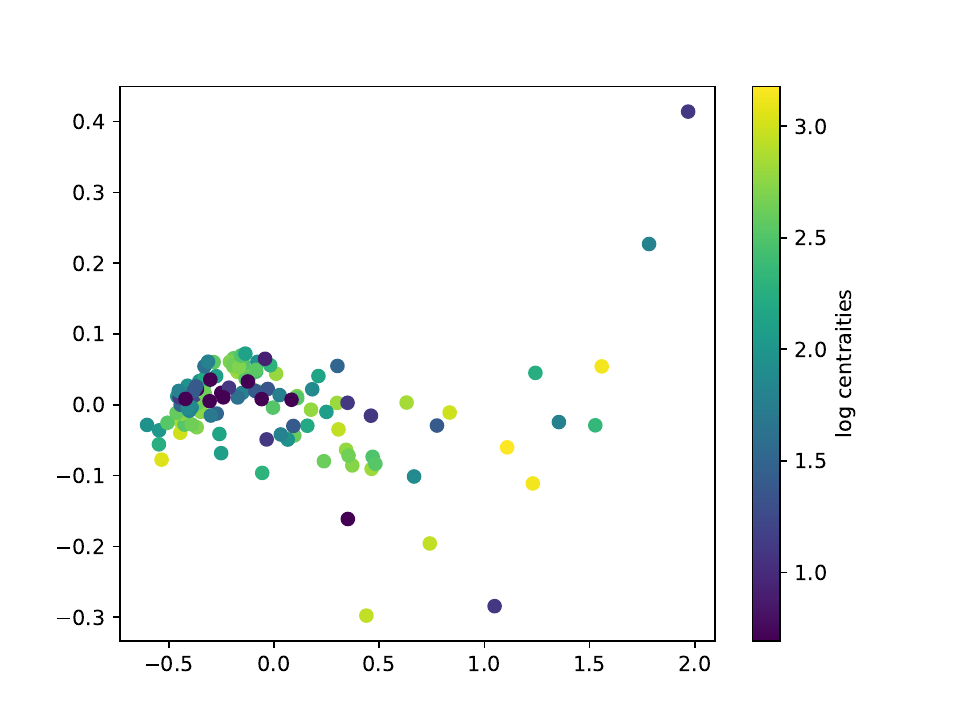}
          \subcaption{Embedding of nodes based on GCN architecture trained for prediction of temporal closeness centrality  in eu-email-dept4}
          \end{subfigure}
            \qquad 
            \begin{subfigure}[c]{.45\textwidth}
              \includegraphics*[width=\textwidth]{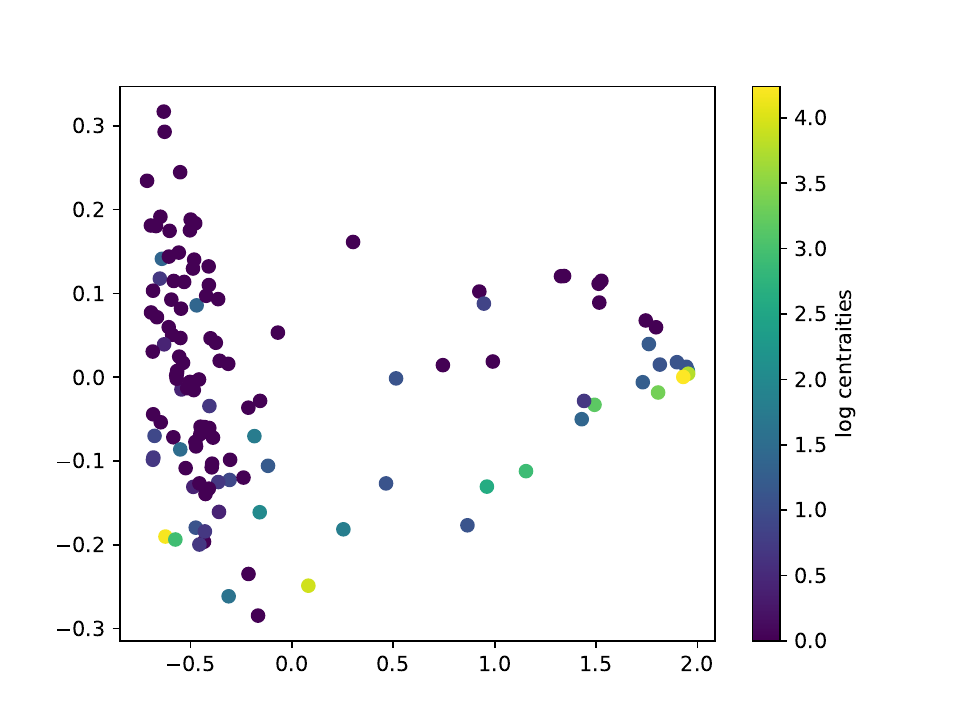}
              \subcaption{Embedding of nodes based on GCN architecture trained for prediction of temporal betweenness centrality in eu-email-dept4}
              \end{subfigure}
  
              \begin{subfigure}[c]{.45\textwidth}
                \includegraphics*[width=\textwidth]{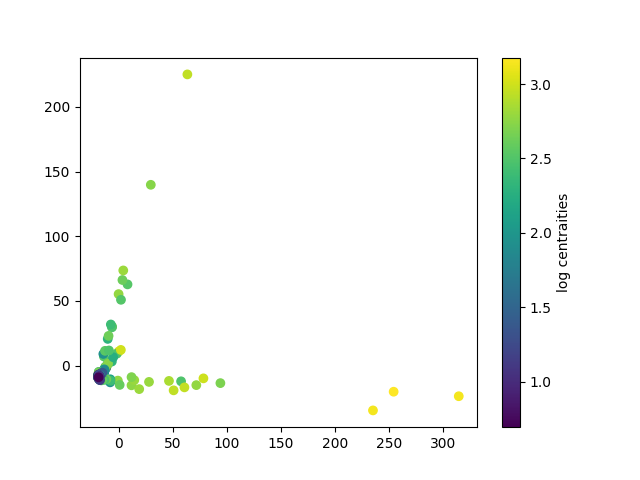}
                \subcaption{Embedding of nodes based on EVO trained for prediction of temporal closeness centrality in eu-email-dept4}
                \end{subfigure}
                  \qquad 
                  \begin{subfigure}[c]{.45\textwidth}
                    \includegraphics*[width=\textwidth]{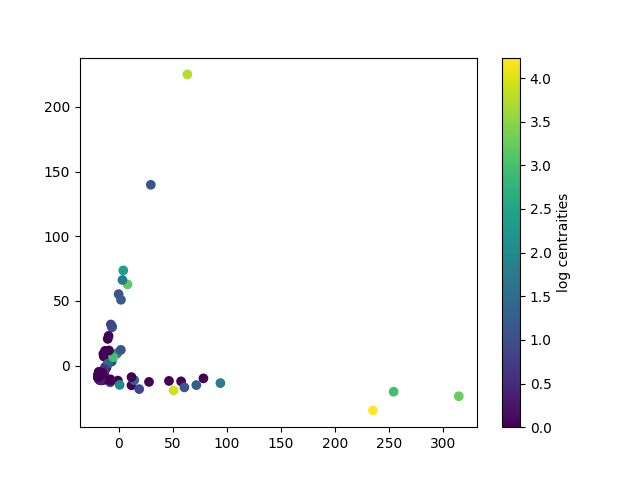}
                    \subcaption{Embedding of nodes based on EVO trained for prediction of temporal betweenness centrality in eu-email-dept4}
                    \end{subfigure}
        \caption{Comparison of node embeddings generated by DBGNN model (top), GCN (middle), and EVO (bottom) for the eu-email-dept4 data set. Nodes are colored according to their temporal closeness (left) and betweenness (right) centrality.}
        \label{fig:embedding}
      \end{figure*}

      The plots in \cref{fig:embedding} show the node emebddings obtained for a GCN and DBGNN model trained to predict temporal closeness centrality (a, b) as well as temporal betweenness centrality (c, d) for the \texttt{eu-email-4} data set. 
  
\FloatBarrier

\end{document}